\documentclass[10pt,journal,cspaper,compsoc]{IEEEtran}

\ifCLASSOPTIONcompsoc
\else
\fi

\ifCLASSINFOpdf
\else
\fi
\pdfoutput=1

\usepackage{subfig}
\usepackage{booktabs}
\usepackage{amsmath}
\usepackage{authblk}

    \usepackage{tikz}
    \usetikzlibrary{arrows}
    \usetikzlibrary{fit}
    \usetikzlibrary{calc}
    \usetikzlibrary{positioning}
    \usepackage{pgfplots}
    \usepackage{xcolor}
    \usepackage{rotating}
    
    \usepackage{longtable}
    \usepackage{graphicx}
    \usepgflibrary{plotmarks}
    \pgfplotsset{compat=newest}

\usepackage{graphicx}
\usepackage{amsmath,amssymb} % define this before the line numbering.
\usepackage{color}
\usepackage[ruled,vlined,linesnumbered]{algorithm2e}
\usepackage{todo,}
\usepackage{verbatim}
\usepackage{booktabs}
\usepackage{multirow}
\usepackage{pdfcomment}
\usepackage[switch]{lineno}
\usepackage{tikz}
\usetikzlibrary{arrows}
\usetikzlibrary{positioning}
\usepackage{pgfplots}
\usepgflibrary{plotmarks}
\pgfplotsset{compat=newest}
\usepackage{lipsum}% http://ctan.org/pkg/lipsum

\newlength\figureheight 
\newlength\figurewidth 
\newlength\mylinewidth

\newcommand{\comments}[1]{}

\newcommand{\mypdfc}[1]{}

\begin{document}

\title{Modeling of Facial Aging and Kinship: A Survey}

%\author{Markos Georgopoulos, Yannis Panagakis and Maja Pantic}

\author[1]{Markos Georgopoulos}
\author[1, 2]{Yannis Panagakis}
\author[1]{Maja Pantic}
\affil[1]{Dept. of Computing, Imperial College London, UK}
\affil[2]{Dept. of Computer Science, Middlesex University London, UK}

\maketitle

\begin{abstract}
Computational facial models that capture properties of facial cues related to aging and kinship increasingly attract the attention of the research community, enabling the development of reliable methods for age progression, age estimation, age-invariant facial characterization, and kinship verification from visual data. In this paper, we review recent advances in modeling of facial aging and kinship. In particular, we provide an up-to date, complete list of  available annotated datasets and an in-depth analysis of geometric,  hand-crafted, and learned facial representations that are used for facial aging and kinship characterization. Moreover, evaluation protocols and metrics are reviewed and notable experimental results for each surveyed task are analyzed. This survey allows us to identify challenges and discuss future research directions for the development of robust facial models in real-world conditions.
\end{abstract}

%% main text
\section{Introduction}\label{S:INTRODUCTION}%

Humans possess explicit, cue-based, and often culturally determined systems for perceiving the facial appearance of their peers \cite{Zebrowitz:2008}. Facial appearance is a primary source of information regarding the person's identity, gender, ethnicity, affective state, head pose, age and  kinship relations. Hence, the perception of facial attributes governs person perception, interpersonal attraction, and consequently prosocial and social behaviour \cite{Alley:1988}, \cite{Park:2008}.

Human face has been thoroughly studied from different, but complementary, perspectives across several disciplines such as neuroscience e.g., \cite{Freiwald:Nature09}, psychology e.g., \cite{Bruce:98}, \cite{Moyse:Age_est_psycho}, sociology e.g., \cite{Exline:1963}, anthropometry e.g, \cite{Farkas:Anthropometry}, medicine e.g., \cite{Dingman:1964} and computer science. From a computational point of view, in particular, advances in computational face modeling enabled the development of reliable methods for automatic detection of faces \cite{Zafeiriou:CVIU15}, recognition  of identity \cite{Zhao:03}, \cite{Jafri:Survey09}, \cite{Abate:2d3dsurvey}, gender \cite{Ng:Gender12}, and ethnicity \cite{Fu:TPAMI14}; detection of salient facial features \cite{Wang:PR17}, \cite{Ekman:77}, estimation of head pose \cite{Ding:ACM16} and analysis of facial expressions \cite{Sariyanidi:TPAMI15}, \cite{Pantic:TPAMI00}, \cite{Zeng:TPAMI09} from visual data. Notably, recently proposed methods match or even achieve better accuracy than humans in several tasks e.g., \cite{Han:TPAMI15}. This progress herald a surge of novel applications in communication, entertainment, cosmetology, and biometrics, to name a few, while facilitating basic research in social sciences and medicine e.g., \cite{Nanni:AIM10}. A thorough list of machine learning and computer vision methods solving the aforementioned face modeling and analysis tasks can be found in the comprehensive survey papers \cite{Zafeiriou:CVIU15}, \cite{Zhao:03}, \cite{Jafri:Survey09}, \cite{Ng:Gender12}, \cite{Fu:TPAMI14}, \cite{Wang:PR17}, \cite{Ding:ACM16}, \cite{Sariyanidi:TPAMI15}, \cite{Zeng:TPAMI09}.

Research towards the development of more detailed computational facial models that capture 
properties of facial cues related to aging and kinship increasingly attracts the attention of the community. Indeed, by capitalizing on recent advances in machine learning, computer vision, and the massive collections of facial data available, significant progress has been made towards addressing the following problems:
\renewcommand{\labelenumi}{\roman{enumi}}
\begin{enumerate}
\item \textit{Age Progression:} that is,   the process of transforming a facial visual input, in order to model it across different ages. The change of the age can be bidirectional, so that the facial output can appear either younger or older than the input. 
\item \textit{Age Estimation:} refers to the process of labelling a facial signal with an age or age group. The input signal can be 2D, 3D or image sequences. The problems that fall into this category can be divided further into two subcategories, depending on the labels of the training data: (a) real age or (b) apparent age estimation, which refers to the age that is inferred by humans based on the individual's appearance.
\item \textit{Age-Invariant Facial Characterization:} involves the process of building a signal representation that is invariant to the facial transformations and appearance changes caused by aging.
\item \textit{Kinship Verification:} is defined as the process of determining whether the individuals in a pair of facial visual inputs are blood related.
\end{enumerate}

Early models for facial age progression and estimation date back to 1994-95 \cite{Kwon:CVPR94}, \cite{Burt:Biological95}, while the problem of face recognition across ages was first investigated in 2000 \cite{Lanitis:FG00}. More recently, since 2010, methods for kinship verification have emerged \cite{Fang:ICIP10}. Since then, the development of 1) robust and computationally efficient models (e.g., AAMs \cite{Cootes:AAM}, CLMs \cite{Cristinacce:CLM} etc) and descriptors (e.g., SIFT \cite{Lowe:SIFT}, HoGs \cite{Dalal:HOG}, LPPs \cite{He:TPAMI05}, SURF \cite{Bay:SURF}, DAISY\cite{Tola:TPAMI10} etc)  of facial appearance, 2) effective machine learning methods such as Boosting \cite{Freund:Boosting} and Support Vector Machines \cite{Vapnik:SVM} and 3) manually annotated facial datasets e.g., MORPH2 (2006), FG-NET (2004), FERET (1998), YGA (2008), Gallagher's Images of Groups (2009), Ni's Web-Collected Images (2009), have facilitated the deployment of reliable computational models for facial aging and kinship. Models, methods, and data for facial aging modeling which have been published before 2010 are thoroughly surveyed in \cite{Ramanathan:JVLC09}, \cite{Fu:TPAMI10}, while an overview of research efforts for facial kinship modeling is currently missing.

This paper aims to provide an up-to-date literature survey of the work done 1) towards the development of facial aging models, complementing previous studies in \cite{Fu:TPAMI10}, \cite{Ramanathan:JVLC09} in several ways and 2) in the emerging topic of facial kinship modeling. Concretely, the aims of this survey are organized as follows:

\begin{itemize}
\item A complete catalogue of publicly available datasets with manual annotations for facial age and kinship modeling tasks is listed in Section \ref{S:data}.  We put particular emphasis on data collected in naturalistic, real-world (in the wild) conditions by providing reviews for 9 recently collected datasets for age modeling and  all the available (i.e., 16) collections of facial images for kinship verification.

\item A comprehensive review of recent as well as seminal methodologies for age progression, age estimation, age-invariant facial characterization, and kinship verification is provided in Sections 4-7. In particular, we provide an in-depth analysis of both geometric,  hand-crafted and learned facial representations for the aforementioned tasks and discuss the type of information they encode as well as their advantages and limitations.  We further elaborate on methods that rely on deep discriminative (e.g., CNNs \cite{Lecun:LeNet}) and generative (e.g., GANs \cite{Goodfellow:NIPS14}) models and appear to be highly effective. Moreover, we review evaluation protocols and metrics, and analyze the most notable experimental results for each surveyed task.

\item The review of data, computational methods for facial aging and kinship reveal useful practices as well as challenges that are yet to be solved. These along with drawn conclusions are discussed in Section 9. 

\end{itemize}

To begin with, a number of modern applications of computational models for facial aging and kinship are discussed in the following section.

\section{Applications}

In this section we present the most significant applications of modeling facial aging as well as kinship in biometrics, forensics, medicine, cosmetology, business and entertainment.

\textit{Biometrics: }The physical, physiological or behavioural cues based on which a person is recognized, e.g., iris, fingerprint, face are referred to as biometrics \cite{Jain:Biometrics}. Age and kinship comprise soft biometrics \cite{Jain:SoftBio}, \cite{Dantcheva:TIFS16}, \cite{Nixon:PRL15} as they can be used to boost the effectiveness of recognition. Besides improving face recognition accuracy there is a need for robustness towards aging and kinship.  Passport checks demand age-invariance in case of large age gap between the passport image and the person in question. Similarly, kinship invariance can potentially boost automatic face recognition, in particular towards distinguishing between kin that look alike.

\textit{Forensics: }Forensics include a set of scientific techniques that are used for crime detection. Among these techniques, forensics art demonstrates the challenging task of producing a lifelike image of a person. In some cases, forensics experts face the need to change the age of a face. Such cases include updating archive images of wanted criminals as well as images of lost children. Additionally, cases such as matching orphaned or lost children and finding the kin of a victim, to name a few, demand the verification of kin relationship of two people. To that end, automatic genealogical research can significantly aid the work of law enforcement agencies.

\textit{Medicine and Cosmetology: }
Being able to model aging and kinship and simulating the transformations on the face is vital for modern medicine and cosmetology. Medical home systems that are used to monitor elderly people can aid medical diagnosis by detecting premature aging. On the other hand, automatic rejuvenation  of the face can serve as a guide for cosmetic surgery. Particularly in the case of children, the parents' craniofacial aging patterns can be used to predict the child's head growth, so that injury-related cosmetic surgery can have optimal long-term results.

\textit{Commercial use: } 
The ever-growing usage of social media and availability of personal photos have led to the rapid integration of facial analysis by businesses. Automatically estimating the customers' age can help with efficient customer profiling and age-oriented decision making, e.g., age-oriented advertisements. Likewise, targeted ads can be more effective when taking kinship into considerations, as people's preferences can be affected by their relatives.

\textit{Entertainment: }Visual effects that age or rejuvenate the actors are already being used in the film making industry. These effects are not limited to movies but are also widely applied to photo editing. The imminent integration of such tools into popular design software will make for more realistic retouche of photos. Make-up artists that specialize in transforming the face can leverage the construction of person, age and kin specific morphable models. Guided by those models, the artists will transform the face of the actor for roles that demand sibling-like similarity between actors.

\section{Datasets}\label{S:data}

The availability of labelled datasets is the cornerstone of the development of facial models for aging and kinship. In this section, we review recently released datasets which contain annotations for age progression, age estimation, age-invariant facial characterization, and kinship verification. Particular emphasis is put on data capturing naturalistic, real-world conditions, often referred to as in the wild \cite{LFW}. A complete catalogue of the available datasets for facial aging modeling is listed in Table~\ref{T:age} while Table~\ref{T:kin} contains the available facial data for kinship verification.

\begin{table*}[htbp]

  \centering
  \caption{Datasets with age labels}
    \begin{tabular}{llllllll}
    \toprule
    Databases & \#data & \#subjects & age-range & precision & In the Wild & Year & modality \\
    \midrule
    
    AgeDB \cite{Moschoglou:CVPRW17} & 16,458 & 568 & 1-101 & exact age & yes   & 2017  & images \\
    IMDB-WIKI \cite{Rothe:IJCV16}& 523,051 & 20,284 & 0-100 & exact age & yes   & 2016  & images \\
    AFAD \cite{Niu:CVPR16}& 164,432 & N/A & 14-40+ & exact age & yes   & 2016  & images \\
    IRIP \cite{Yang:TIP16}  & 2,100 &   N/A    & 1-70  & exact age &  yes     & 2016  & images \\
    OUI-Adience \cite{Eidinger:TIFS14}& 26,580 & 2,284 & 0-60+ & age group & yes   & 2014  & images \\
    CACD \cite{Chen:TransMult15}& 163,446 & 2,000 & N/A    & exact age & yes   & 2014  & images \\
    UvA-Nemo \cite{Dibekliouglu:ECCV12}& 1,247    & 400     & 8-76 & exact age & no   & 2014  & videos \\
    VADANA \cite{Somanath:ECCV12}& 2,298 & 43    & 0-78  & age group & yes   & 2011  & images \\
    LHI Face Dataset \cite{LHI}& 8,000 & 8,000 & 9-89  & exact age & no    & 2010  & images \\
    HOIP \cite{HOIP}& 306,600 & 300   & 15-64 & age group & no    & 2010  & images \\
    FACES \cite{Ebner:2010}&  1026 & 171  & 19-80 &  exact age&  no   & 2010  & images \\
    Web Image Db \cite{Ni:ACM09}& 219,892 & N/A   & 1-80  & exact age & yes   & 2009  & images \\
    Images of Groups \cite{Gallagher:Groups}& 28,231 & 28,231 & 0-66+ & age group & yes   & 2009 & images \\
    YGA \cite{Guo:TIP08}& 8,000 & 1,600 & 0-93  & exact age & yes   & 2008 & images \\
    Iranian Face \cite{IranianFace}& 3,600 & 616 & 2-85  & exact age & yes   & 2007 & images \\
    Brown Sisters \cite{BrownSisters}& 16    & 4     & 15-62 & exact age & yes   & 2007  & images \\
    Scherbaum's Db \cite{Scherbaum:CGF07}& 438 & 438 & 8-18 or adult & both & no   & 2007 & 3D\\
    MORPH 2 \cite{Morph}& 55.134 & 13.618 & 16-77 & exact age & no    & 2006 & images \\
    WIT-BD \cite{witbd}& 26,222 & 5,500 & 3-85  & age group & yes   & 2006  & images \\
    AI\&R \cite{fu:air}& 34 & 17 & 22-66 & exact age & no   & 2006 & images \\
    FRGC \cite{frgc}& 50,000  & 568 & 18-70 & exact age & partially & 2005 & images, 3D \\
    Lifespan Database \cite{Lifespan}& 576 & 576 & 18-93 & age group & yes   & 2004 & images \\
    FG-NET \cite{FGNET},\cite{Lanitis:TPAMI02prog}& 1.002 & 82 & 0-69  & exact age & yes   & 2004 & images \\
    PIE \cite{pie}& 41,638 & 68    & N/A    & exact age & no    & 2002  & images \\
    FERET \cite{feret}& 14,126 & 1,199& N/A    & exact age & partially & 1998 & images \\
    Caucasian Face Db \cite{Burt:Caucasian}& 147   & 147     & 20-62 & exact age & no    & 1995  & images \\
    \bottomrule
    \end{tabular}%
  \label{T:age}%

\end{table*}%

\begin{table*}[htbp]

  \centering
  \caption{Datasets with kinship labels}
    \begin{tabular}{llllllll}
    \toprule
     Databases & \#data & \#kin pairs & \#relationships & In the Wild & same photo & Year  & modality \\
     \midrule
    KFVW \cite{Yan:PR17} & 836   & 418  & 4     & yes   &  N/A  & 2017  & videos \\
    WVU Kinship Db \cite{Kohli:TIP16} & 904   & 113  & 7     & yes   &  N/A  & 2017  & images \\
    Families In the Wild \cite{FIW}&   11,193 & 418,060 &   11   &   yes &   partially & 2016  & images\\
    TSKinface \cite{Qin:TransMulti15}& 787   & 1,015x2 & 4     & yes   & yes   & 2015  & images \\
    Sibling-Face \cite{Guo:ICPR14}&   N/A &  720&     3 &   yes &    no&   2014& images\\
    Group-Face \cite{Guo:ICPR14}&    106&  395&     7 &   yes &   yes & 2014  & images\\
    HQfaces \cite{Vieira:VisCom14} & 184   & 92   & 3     & no   & no    & 2014  & images \\
    LQfaces \cite{Vieira:VisCom14} & 196   & 98   & 3     & yes   & no    & 2014  & images \\
    KinFaceW-I \cite{Lu:TPAMI14}& 1,066 & 533   & 4     & yes   & partially & 2014 & images \\
    KinFaceW-II \cite{Lu:TPAMI14}& 2,000 & 1,000 & 4     & yes   & yes   & 2014  & images \\
    Family 101 \cite{Fang:ICIP13}& 14,816 & 206   & 4     & yes   & no    & 2013  & images \\
    IIITD-Kinship \cite{Kohli:BTAS12} & 544 & 272   & 7     & yes   & N/A & 2012 & images \\
    VADANA \cite{Somanath:ECCV12}& 2,298 & 69 & 7     & yes   & yes   & 2012  & images \\
    UvA-Nemo \cite{Dibekliouglu:ECCV12}& 515   & 95    & 7     & no    & no    & 2012  & videos \\
    UBKinface \cite{Xia:IJCAI11}, \cite{Shao:CVPRw11}& 600   & 400   & 4     & yes   & no    & 2011  & images \\
    CornellKin \cite{Fang:ICIP10}& 300   & 150   & 4     & yes   & partially & 2010  & images \\
    \bottomrule
    
    \end{tabular}%
  \label{T:kin}%

\end{table*}%

\subsection{Datasets with age labels}

The vast majority of the available datasets for facial aging modeling contain still images and apart from the FACES, IRIP, LHI, and YGA face datasets, they are not balanced with regards to the gender and age of the subjects, as shown in Table \ref{T:age}. Also, while containing an abundance of different annotated faces, many of these datasets do not contain a considerable number of images of the same person at different ages, which is essential for training methods for age progression and age-invariant facial characterization.

For all aging modeling tasks, the most widely used benchmarks, and their respective aging databases, are FG-NET and MORPH (alboum 2). The \textit{Face and Gesture recognition Network database (FG-NET)} contains 1,002 images of 82 subjects, which are mostly Caucasians. The age of the subjects ranges from newborns to 69 years old, while the gap between the age of the same person in different images ranges from 0 to 54 years. This dataset was collected from personal photo albums and contains coordinates of facial landmarks for each image. The \textit{MORPH} dataset was released in 2006 and its second album contains 55,134 images of 13,618 people. The subjects are mostly African while there are $4$ images of each person on average.

\subsubsection{Data suitable for age estimation}

The databases used for the task of age estimation vary greatly in terms of sample size, number of subjects,  and age-range, as indicated in Table~\ref{T:age}. Nevertheless, the recent success of deep learning-based models in computer vision has created a need for larger datasets with age annotations. Towards this end, the IMDB-WIKI, AFAD and OUI-Adience datasets have been collected.

\textit{OUI-Adience \cite{Eidinger:TIFS14}:}
The OUI-Adience dataset contains 26,580 facial images from the albums in the website Flickr.com. These photos were made publicly available through the Creative Commons license. The photos were collected from approximately 200 albums, while the Viola-Jones \cite{ViolaJones}  detector was used to detect the faces. The age-range of the dataset is 0-60+, although the ages above 48 years old are less represented.

\textit{UvA-NEMO \cite{Dibekliouglu:ECCV12}:} The UvA-NEMO dataset consists of 1,240 videos of 400 people smiling. The smiles are both posed and spontaneous and were captured under controlled conditions. The age of the faces varies from 8 to 76 years.

\textit{AFAD \cite{Niu:CVPR16}:}
The Asian Face Age Dataset (AFAD) was introduced by Niu et al. and contains 164,432 face images. These images were collected from the RenRen Social Network which is widely used by Asian students. Therefore, most subjects are Asian and under 30 years old. The date of birth of every subject was provided by the respective user account.

\textit{IMDB-WIKI \cite{Rothe:IJCV16}:}
The largest age-annotated dataset is the IMDB-WIKI dataset. The dataset consist of 523,051 facial images that were crawled from the Wikipedia and IMDB websites. All metadata were also collected from the above mentioned web-sites.  The average age of the dataset is about 32 years old.

The age annotations to all the datasets in Table~\ref{T:age} refer to real age, and therefore cannot be used for the task of apparent age estimation. In order to address the lack of such datasets Escalera et al. \cite{Escalera:ICCVw15} built the Chalearn-AgeGuess dataset, which is the first dataset with apparent age annotations. This dataset contains 4,691 images and was used for the Chalearn  competition. The annotations were acquired using the AgeGuess online voting platform. Using the same platform, as well as Mechanical Turk workers, the APPA-REAL \cite{Agustsson:APPAREAL} dataset was introduced by Agustsson et al. This dataset contains 7,591 faces with real and apparent age annotations in more than 7,000 images. The age range is between 0 and 95 years.

\subsubsection{Datasets for Age Progression and Age-Invariant Facial Characterization}

The tasks of age progression and age-invariant facial characterization require data containing the same person at different ages. Therefore, the collection of such databases is a challenging procedure. Besides the FG-NET and MORPH datasets, the recently created AgeDB, VADANA and CACD datasets are also widely used for age progression and age-invariant facial characterization tasks. A brief overview of these datasets is presented below.

\textit{AgeDB \cite{Moschoglou:CVPRW17}:}
AgeDB contains 16,488 images of 568 subjects, manually collected from Google Images. Avoiding to collect the images in a semi-automatic manner ensures that the labels are not noisy. Each subject in the dataset has 29 photos on average, while the average age-range of the subjects is 50.3 years.

\textit{VADANA \cite{Somanath:ECCV12}:}
The VADANA dataset contains 2,293 images from only 43 subjects (26 males and 17 females) that are mostly South-Asian. It contains 3-300 images per subject with wide variation in pose, illumination and occlusions, while the maximum age gap between two images of the same subject is 37 years. The annotations of this dataset include amongst other attributes, kinship and occlusions (e.g., facial hair and glasses).

\textit{CACD \cite{Chen:TransMult15}:}
The Cross-Age Celebrity Dataset (CACD) was published in 2014 and contains 163,496 images from 2,000 celebrities. It was collected from Google Images using information from IMDB.  The dataset contains more than 80 images per subject on average, but the maximum age gap between these images is 10 years. This is due to the fact that the collected images were captured between 2004 and 2013. The subjects are mostly Caucasian.

Table \ref{T:age} reveals that age modeling has been primarily studied by employing datasets captured in the wild. The first major benchmark in the wild is FG-NET and traces back to 2004. Ever since, the majority of publicly available datasets are captured under unconstrained conditions, albeit having a small number of images. Images from the FG-NET or Morph2 datasets exhibit limited pose and image quality variation. These issues deteriorate the real-world performance of age modeling systems, especially when dealing with extreme poses or small and blurry images. The recent advances in face detection technology \cite{Mathias:ECCV14}, \cite{Zhang:SPL16} has resulted in much larger automatically collected datasets e.g., IMDB-WIKI, CACD, OUI-Adience. These datasets contain a greater number of images and exhibit greater variation with regards to pose, expression and image quality. The reported baseline performance for the OUI-Adience is much lower (i.e.,  classification accuracy of 45.6\%) than older datasets, which is characteristic of the challenging nature of the benchmark.

\subsection{Datasets with kinship labels}

Kinship verification from faces is a relatively new task,
aiming to verify whether the individuals in a pair of facial visual inputs are blood related. There are eleven different kinds of blood relations: (i) Father- Daughter, (ii) Mother- Daughter, (iii) Father- Son, (iv) Mother- Son, (v) Sister- Sister, (vi) Sister- Brother, (vii) Brother- Brother. (viii) Grandfather- Grandson, (ix) Grandmother- Grandson, (x) Grandfather- Granddaughter and (xi) Grandmother- Granddaughter.
A catalogue of dataset with kinship annotations is tabulated in Table \ref{T:kin}. The vast majority of the available data is annotated in terms of only 4 kin relations.

\textit{CornellKin \cite{Fang:ICIP10}:}
The first widely used dataset with kinship annotation is the CornellKin Database. The dataset consists of 300 images and 150 kin pairs, gathered through online searches.

\textit{UB KinFace \cite{Xia:IJCAI11}, \cite{Shao:CVPRw11}:}
In 2011, Shao et al. published the UB KinFace Ver 2.0 dataset, which was an extension to UB KinFace Ver 1.0. The UB KinFace consists of 600 web-collected images and 400 kin pairs, half of which are Asian. Every kin pair is represented by an image of the child, an image of the parent at a young age and an image of the parent at an older age.

\textit{IIITD Kinship Database \cite{Kohli:BTAS12}:}
The IIITD Kinship Database was released in 2012 and contains 544 images of 272 kin pairs in unconstrained environment. The dataset also includes 272 non-kin pairs. The faces are mainly Indian and American, while other ethinicities include Asian and Afro-American. The dataset contains 7 kin relations.

\textit{Family 101 \cite{Fang:ICIP13}:}
The Family 101 dataset was published in 2013 and consists of 14,816 images of 607 people, collected from Amazon Turk workers. It is also the first one to indicate the structure of the families. In particular, the dataset contains whole family trees, each containing 1 to 7 nuclear families. There are 206 families, each having 3 to 9 family members. The subjects depicted in the data are mostly Caucasian.

\textit{KinFaceW I \& II \cite{Lu:TPAMI14}:}
The most widely used kinship-annotated datasets are the KinFaceW I \& II albums that were intoduced by Lu et al. in 2012. The KinFaceW I album contains 1,066 images and 533 kin pairs, while the KinFaceW II contains 2,000 images of 1,000 pairs. All the images are collected from the web, while the difference between the two albums is that KinFaceW II contains kin-related faces cropped from the same image. 

\textit{HQ \& LQ Faces \cite{Vieira:VisCom14}:} In order to investigate  the ability to generalize from high-quality images captured under controlled condition to low-quality images gathered from the Internet, Vieira et al. introduced the HQfaces and LQfaces datasets. The HQfaces dataset contains frontal images of 92 sibling pairs captured by a professional photographer in neutral expression. The LQfaces dataset consists of 98 sibling pairs collected from the Internet with varying resolution. The subjects in the two datasets are mainly Caucasian.

\textit{Sibling-Face \& Group-Face \cite{Guo:ICPR14}:}
In 2014, motivated by the lack of large number of sibling pairs, Guo et al. introduced the Sibling-Face dataset. The images were collected from sites like Flickr, while the annotations were obtained from the tags and descriptions of the images. Along with this dataset, the Group-Face dataset was also introduced, containing images of groups of kin-related people.

\textit{TSKinface \cite{Qin:TransMulti15}:}
Qin et al. introduced the TSKinface dataset, which consists of 787 images and 1015 family groups. These groups contain the father, the mother and child or children. The are 274 photos of Father-Mother-Daughter families, 285 Father-Mother-Son families and 228 Father-Mother-Daughter-Son families. The majority of the subjects included in this dataset are Asian.

\textit{Families in the Wild \cite{Robinson:FIW}:}
One of the most recent efforts in kinship annotated datasets is the Families in the Wild  (FiW) dataset, containing 11,193 images. The FiW dataset is the largest kinship annotated dataset to date, containing 30,725 face-images of 10,676 individuals. The dataset includes 11,193 family photos, while it is the first to have 2,060 annotated pairs of grandparents and grandchildren.

\textit{WVU Kinship Database \cite{Kohli:TIP16}:} The WVU Kinship Database was introduced in 2017 by Kohli et al. and contains 113 kin-related pairs. The dataset allows for intra-class variation, as it contains 4 images per person. The dataset is not balanced and consists of: 22 Brother-Brother pairs, 9 Brother- Sister pairs, 12 Sister -Sister pairs, 12 Father- Daughter pairs, 34 Father- Son pairs, 12 Mother- Daughter pairs and 8 Mother- Son pairs.

\textit{Kinship Face Videos in the Wild Dataset \cite{Yan:PR17}: } The Kinship Face Videos in the Wild (KFVW) dataset contains 418 face videos from TV shows. Each video is 100-500 frames long and presents great variation in pose, occlusions and expressions. The baseline experiments on the dataset achieved poor performance and thus, the authors note the challenges of advancing video-based kisnhip verification with this dataset.

Similar to age related tasks, the problem of kinship modeling has been attacked using mainly in the wild datasets. The nature of the labels makes the task of gathering and annotating such datasets very laborious. As a result most of the datasets in Table \ref{T:kin} contain a small number of images, which is not ideal for modeling large variations. What is more, these images, more often than not, originate from the same photo. Cropping the data from the same image can significantly bias the task of classification, by adding factors like the environment, lighting, chrominance and the image quality \cite{Lopez:TPAMI16}. Therefore, datasets that contain such images are considered biased and can not be used for the task of kinship verification. This is evident when comparing the state-of-the-art performance between classic benchmarks like KinFaceW-II (i.e., classification accuracy 96.2\%) and more recent ones like FIW (i.e., classification accuracy 71\%).

It is worth mentioning that, VADANA and UvA-NEMO Smile Databases contain both kinship and age annotation. In particular, for the Smile Dataset, the kinship annotations were obtained based on the names of the subjects. 

In the following sections, both seminal and recent methods for age progression, age estimation, age-invariant facial characterization, and kinship verification are reviewed. Particular emphasis is put on the facial representations employed. That is, for each of the aforementioned tasks, the presented methods are grouped according to the type of information captured by the employed facial representations. Broadly, geometric,  hand-crafted, and learned facial representations are considered. Discussion on corresponding classification and regression methods is also provided. Furthermore, for each task evaluation protocols and metrics are reviewed, and the most notable experimental results are analyzed.

\section{Age estimation}

A significant volume of research has been done in  age estimation from facial visual data. Since the labels of the various age-annotated datasets are either discrete ages or correspond to age-groups,
the problem can be naturally cast as a multiclass classification problem, where the classes represent discrete ages or age-groups. However,  neighbouring labels (ages, or age groups) might share important information which is neglected by  classification methods. This is addressed by regression methods which appear to perform better. Nevertheless, appearance changes more rapidly during youth and slower in adults. To alleviate this, non-stationary kernels can be employed; yet learning with kernels is prone to overfitting. A different approach to deal with this challenge is to adopt ranking methods e.g., \cite{Chang:ICPR10},\cite{Chang:CVPR11}, that learn an individual classifier for each age class.
 In the following subsections several methods for age estimation are reviewed. As already mentioned, they are organized according to the type of  facial representation employed.

\subsection{Geometric facial representation-based methods} 

 The tasks of automatic age estimation  from facial images introduced in 1994 by  Kwon and Lobo \cite{Kwon:CVPR94}.
Inspired by anthropometric studies \cite{Alley:1988} that describe the growth of the human head from infancy to adulthood, six facial distance ratios are used to discriminate between infants and adults. The adult faces are further classified into young adults and older adults by using snakelets (i.e., deformable curves) \cite{Kass:IJCV88}, which  capture wrinkles on certain regions. 

The use of such anthropometric models, that represent the shape of the human head, introduces a number of challenges for age estimation. For instance,  geometric descriptors are
 sensitive to pose variations and can therefore be applied only to frontal images. Several methods deal with this challenge by normalizing the faces via Procrustes analysis \cite{Kendall:84}. Invariant to small changes in camera location and head pose geometric facial representation has been proposed in  \cite{Wu:TIFS12} and \cite{Thukral:ICASSP12} by employing the properties of Grassmann manifold \cite{Edelman:98}. Age estimation is obtained via Support Vector Machine (SVM)-based regression in \cite{Wu:TIFS12} and 
 Relevance Vector Machine (RVM) \cite{Tipping:JMLR01} in
 \cite{Thukral:ICASSP12}.
 
Experimental results on the FG-NET dataset in \cite{Wu:TIFS12} indicate that even though geometric information is sufficient for age estimation in young ages, texture information is needed for accurate age estimation in adults. To alleviate this, 
fusion of geometric and texture information (represented by Gabor phase patterns (HGPP)) is applied \cite{Zhang:TIP07}.

\subsection{Appearance-based methods}

Appearance models allow to capture texture information of the face along with its shape. A wide variety of appearance models have been proposed in the literature. Below, we present those that have been extensively used in the context of age estimation. 

\textit{Active Appearance Models:}
Active Appearance Model (AAM) is a generative facial model introduced in \cite{Cootes:AAM} by Cootes et al. AAMs employ Principal Component Analysis (PCA) to learn a linear model for shape and appearance from images and a set of landmarks. This representation was first used for age estimation in \cite{Lanitis:TPAMI02}. In order to capture 95 per cent of the data variation, 50 parameters of AAM are used and age estimation is formulated as a regression problem in the parameter space. Besides age estimation, the effectiveness of the model in simulating the aging process is evaluated on age progression as well as age-invariant face recognition.

Most of the datasets in Table \ref{T:age} are imbalanced; that is, the labels of the data are sparse and not evenly distributed. To overcome this problem, the age labels are encoded in a cumulative manner in \cite{Chen:CVPR13} using Cumulative attribute (CA) vectors. If the face is older than the \textit{i}th age, the \textit{i}th element of its CA vector is \textit{1} and \textit{0} otherwise. In this way, data points with similar labels have a similar cumulative attribute vector. The CA is obtained from the parameters of AAM and Support Vector Regression (SVR) is performed in the CA space. The experiments on FG-NET indicate that shape features play a vital role in age estimation of young faces, while texture features become increasingly discriminative after the age of 20.

In order to make use of the ordinal information of the age labels, Chang et al. \cite{Chang:ICPR10}, \cite{Chang:CVPR11} introduced a ranking approach for the task of age estimation in 2010 using AAMs. A ranking model gradually splits the feature space as a binary classification problem is solved for each age label \textit{k}. The splits are produced by conducting a query '\textit{is this face older than age k?}'. The final estimation is then produced by a ranking rule based on the outputs of the classifiers. Nevertheless, learning a large number of classifiers can be computationally intensive. More recent ranking approaches introduce AAM-based methodologies to reduce the computational cost, e.g., \cite{Liu:ICASSP16}, \cite{Liu:ICIP16}. In particular, a Partial Least Squared-based Ranker, which learns all classifiers jointly, is introduced in \cite{Liu:ICASSP16}. Also, a more time-efficient algorithm is proposed in \cite{Liu:ICIP16}. A Linear Canonical Correlation Analysis-based Ranker is employed and competitive results are obtained with a lower number of parameters. The smaller complexity and model size make such approaches ideal for real-time applications.

In the above papers the effectiveness of the texture and shape parameters of AAM as a facial representation for the task of age estimation have been demonstrated. However, AAM uses PCA and can therefore only account for linear modes of variation, which introduces a set of challenges when generalizing to unseen data. In order to capture non-linear variations such as expressions and poses, Duong et al. \cite{Duong:CVPR15} introduced the Deep Appearance Model (DAM). This generative model is based on Deep Boltzman Machines \cite{Sala:DBM}, a hierarchical method that can capture variations in shape and texture that could be higher than second order. Experiments in the FG-NET dataset indicate the ability of DAM to outperform AAM, while being more robust to additive noise. Nonetheless, having more parameters, this method is more resource intensive in comparison to traditional AAM. Deep hierarchical representations are analyzed further in subsection \ref{subS:deep}.

\textit{Local Binary Pattern:}
The Local Binary Patterns Histogram (LBP) \cite{Ojala:LBP} is an effective texture descriptor which has been widely employed for
texture classification and face recognition \cite{Ahonen:TPAMI06}. LBPs are  binary codes computed in the neighbourhood of pixels and their histogram is obtained for different image-patches. Some variations of the descriptor \cite{Wolf:ECCVw08} compare the values of three (Three Patch-LBP) or four (Four Patch-LBP) patches to produce a code for each pixel. Yang et al. \cite{Yang:Bio07} employed this facial representation to classify faces into three age-classes. In particular, LBP Histograms are extracted from a large set of possible face patches and  AdaBoost \cite{Schapire:99} is applied next to choose the most discriminative ones. Several different regression methods are applied to the extracted representations, exhibiting promising results. In \cite{Eidinger:TIFS14}, fusion of LBP and FP-LBP \cite{Wolf:ECCVw08} is exploited for age group and gender classification via SVM  classifier.

\textit{Wavelet-based Features:} 
Wavelet-based features have been widely used in facial analysis, due to their ability to robustly capture texture information. Among the most widely used type of wavelet in age estimation is the Haar wavelet \cite{Mallat:Wavelets}, yielding features that are robust to appearance variation. Haar-like features are employed for the task of age estimation in \cite{Zhou:ICCV05}. Age estimation is posed as a regression problem with a regularized $L_2$ loss function using boosting.

Gabor wavelet has also been widely applied in texture analysis \cite{Liu:TIP02}. Importantly, this wavelet has a biological significance, since their kernels are similar to the receptive field of the mammalian cortical simple cells. The magnitude of the coefficients, which is obtained by convolving facial images with Gabor wavelets across different scales and orientations, is used as facial representation in  \cite{Gao:ICB09}.  Age-group classification is performed using Fuzzy LDA. Experiments on a private dataset indicate improved performance compared to that obtained by employing LBPs.

To capture curvature information, like wrinkles, extended curvature Gabor (ECG) filters are used in \cite{Kim:ICIP15}. After feature selection, Random Forest (RF) Regression is used for age estimation. Another interesting wavelet based representation  is the scattering transform (ST) \cite{Mallat:GroupScatter}. ST is calculated by cascading wavelet modulus operators along different paths in a deep convolution network. A number of recent works use Gabor wavelets with ST for age estimation. A ranking approach using ST representations is introduced in \cite{Chang:TIP15}. The experiments reveal that AAMs performs better than ST in FG-NET dataset, while the opposite stands for the MORPH2 dataset. This is attributed to the fact that, FG-NET has significantly more images per person and fewer people than MORPH2, capturing long-term person-specific age information. Hence, the AAM can efficiently model the intra-person variance, while the ST appears to model person-invariant age information better.

\textit{Biologically Inspired Features:}
Besides the biologically relevant Gabor wavelets, the Biologically Inspired Features (BIF) \cite{Guo:CVPR09} have also been proposed as suitable facial representation for  age estimation. Inspired by the HMAX model \cite{HMAX}, the BIF pipeline consists of simple (S1) and complex (C1) layers imitating  the classical Hubel and Wiesel model \cite{HubelWiesel} for the primary virtual cortex. Gabor filters are used for S1 units, while the C1 units pool over the S1 units with a non-linear maximum operator 'MAX' and normalizes with a standard deviation operator "STD". It is argued that the "STD" operator reveals local variations capturing vital aging information, like wrinkles and eyelid bags. Experiment are conducted on the FG-NET and YGA datasets using SVM classification and SVR regression. Experimental results indicate that classification accuracy drops  on small imbalanced datasets like the FG-NET.

A variety of different regression methods have been proposed using the BIF descriptor. In particular, a non-linear extension of Partial Least Squares (PLS) \cite{Geladi:PLS} is introduced in \cite{Guo:CVPR11}. Kernel PLS (KPLS) regression is used on BIF feature space for simultaneous dimensionality reduction and function learning. Another BIF-based method, particularly an age estimator that is robust to pose variation, is introduced in \cite{Song:ICCV11}. The multi-view property of this method is obtained by using video information in a semi-supervised manner. That is, a regression problem is solved with a regularization factor that enforces output consistency throughout an unlabelled video sequence.

A fusion of classifiers and regressors is used in \cite{Han:TPAMI15}, where an automatic demographic estimator is introduced. A hierarchical fusion of SVM classifiers and SVR regression is used for age estimation. That is, a face is passed through a series of binary classifiers, indicating different age groups in a coarse-to-fine manner, before going through an SVR regressor. A comparative study between the automated system and human workers is also presented in \cite{Han:TPAMI15}, revealing the inability of the humans to perform accurate real age estimation, as well as the challenges of crowd-sourcing such efforts.

A set of regression methods that take into account the correlation between age classes is proposed in \cite{Geng:TPAMI13}, \cite{Geng:ICPR14}. Based on the fact that aging is a slow and smooth process, the information extracted from a face does not only describe the exact age of the subject, but also the neighbouring ages. Therefore, based on that assumption the model assigns a label distribution to the data \textit{p(y$\mid$x)}, instead of the exact label. For that purpose two label distribution learning algorithms, namely IIS-LLD \cite{Geng:TPAMI13} and CPNN \cite{Geng:TPAMI13}, are proposed and tested in FG-NET and MORPH2 datasets, using AAMs and BIFs respectively.

\subsection{Subspace learning-based methods}

The aforementioned facial representations are extracted from  each and every image individually. Nevertheless, the images of the same person at different ages are correlated and form a pattern to be recognized. To investigate this concept, the AGing pattErn Subspace (AGES) method is introduced in \cite{Geng:TPAMI07}. In order to represent the transformations caused by aging, aging patterns are employed. These patterns are obtained as sequences of face images sorted in time order. Each individual in the training set has a different aging pattern, which is obtained from the training images and by solving a missing value problem for the missing ages. By applying PCA on the resulting aging pattern vectors, the so-called Aging Pattern Subspace is obtained. Each point on this subspace represents an aging pattern. During testing, an unseen face is projected on the aging pattern subspace and an aging pattern is selected based on the reconstruction error. That is, the input face is projected on all the possible positions in each aging pattern and the generated face is compared to the input. The selected position on the selected aging pattern is then returned as the estimated age. Instead of pixel intensities, the method uses low-level face representation extracted from a different model, in particular the AAM.

One of the shortcoming of the AGES method is that it requires a large number of images of the same person at different ages to obtain the aging patterns. In practice, most datasets are very sparse, containing only a few images per person that do not span a large range of ages. To overcome this challenge, Fu et al. \cite{Fu:TransMulti08} uses face images from all the people in the training set to obtain the aging pattern. In that way, the common underlying aging pattern of the data is captured. By representing each age label with multiple images, a discriminative manifold \cite{Seung:Manifold} is obtained in a supervised manner. During testing, the input face is embedded on this low dimensional subspace and a regression problem is solved. Several methods are applied to visualize the geometry of the manifold, e.g., Locality Preserving Projections (LPP) \cite{He:TPAMI05}, Orthogonal Locality Preserving Projections (OLPP) \cite{Cai:TIP06} and Conformal Embedding Analysis (CEA) \cite{Fu:CVPR07}, with the results indicating a distinct pattern. This is contrary to results produced by unsupervised methods, e.g., PCA, that do not take into account the age information. Different regression methods have been applied on the aging manifold. Particularly, simple regression methods, i.e. linear, quadratic and cubic, are used in \cite{Fu:TransMulti08}. A more sophisticated method is proposed in \cite{Guo:TIP08}, where Locally Adjusted Robust Regression (LARR) is introduced.

The aging manifold on the aforementioned methods spans the image space. That is,  pixel intensities are used as local descriptors. This can deteriorate the discriminative ability of the model, since the image space represents all variations and is not particularly age-sensitive. In order to incorporate the texture information in the aging manifold, LBP are used as low-level representations in \cite{Fang:ICPR10} where the  age is obtained using regression methods, e.g., a Neural Networks (NN) and Quadratic Funtion (QF). Similarly, aging manifold can be obtained from different descriptors, e.g., Gabor \cite{Li:CVPR12} and BIF \cite{Li:TransCyber15}. Particularly, an aging manifold that preserves the ordinal information as well as the geometric structure of the data is obtained in \cite{Li:CVPR12}, \cite{Li:TransCyber15}. The resulting subspace is well-suited for ranking methods and therefore the OHRank is employed for age estimation.

Most manifold learning algorithms assume that the feature space is locally Euclidean and therefore use a Euclidean metric to determine neighbourhoods. In the aging manifod, this is not always the case, due to the non-linear nature of aging. Therefore, a distance metric adjustment to LPP is introduced in \cite{Chao:PR13}. The new metric is learned using Label-sensitive Relevant Component Analysis (RCA) \cite{Bar:ICML03}. Ordinal information is also incorporated in the aging manifold and k-Nearest Neighbours (kNN) and SVR are used for age estimation on the embedded subspace. Similarly, a subspace with ordinal information is captured in \cite{Chen:TIFS13}. Pairwise age ranking is utilized to obtain a subspace that minimizes the distance between images of the same label, under a set of ranking constraints.

\subsection{Deep learning-based methods} \label{subS:deep}

Among the different sources of facial variation, the non-linear transformations, such as deformations due to expressions, pose and age, are the most challenging to model. To model such non-linear variations, some of the aforementioned methods extract simpler features at different levels in a hierarchical manner. Since deep architectures build non-linearities on top of each other, they can learn non-linear transformations more efficiently than simple models and perform better in most tasks, albeit being more vulnerable to overfitting. A classical deep architecture is the feed-forward Artificial Neural Network (ANN). This model uses pixel intensities (PI) or low-level descriptors as inputs and builds a subspace in a supervised manner using backpropagation. A variation of ANN, namely the Compositional Pattern Producing Network (CPPN) \cite{Stanley:CPNN}, is used in \cite{Fan:ICCV11} for the task of age estimation. Contrary to typical ANNs, the four-layer ANN proposed in \cite{Fan:ICCV11} has different transfer functions for each neuron. 

Similar to the ANNs,  Convolutional Neural Networks (CNN) \cite{Lecun:LeNet} are used to extract high-level features, or perform regression or classification as an end-to-end model. Instead of using handcrafted features, a CNN applies filters on the raw images in a hierarchical supervised manner. Convolution and subsampling are applied iteratively to create each layer's feature map. The filters are then optimized using backpropagation, while the architectures used can vary greatly. Deep CNNs have outperformed classical methods in most tasks, including age estimation. Yang et al. \cite{Yang:CVPR11} is the first to use a 5-layer CNN for age estimation. The model has multiple outputs for gender, age and race, while the age estimation performance is worse than that obtained by BIF. In order to improve the performance, Yi et al. \cite{Yi:ACCV14} adopts methodologies from traditional facial analysis and used a fully connected ensemble of 23 CNNs. Each CNN is applied on different face patches and the method succeeds in improving the state-of-the-art performance on the MORPH 2 dataset.

Recently, the large availability of data and computational power have boosted the popularity of deeper and more complex architectures (e.g., ResNet \cite{He:ResNet}, DenseNet \cite{Huang:DenseMet}). A hybrid deep multi-task CNN is employed in \cite{Xing:PR17} to predict age, race and gender. This architecture incorporates the gender and race predictions by training a different age estimator for each group. Another deep CNN architecture that leverages weakly labelled data is proposed in \cite{Hu:TIP17}. The proposed model performs age estimation with assistance of age difference information (AEAD). The age difference information is obtained from image pairs of the same subject with known age gap.

Deep ranking methods have found great success in age estimation. A deep ordinal regressor is trained on the very large AFAD dataset in  \cite{Niu:CVPR16}. This ranking approach uses multiple output CNNs (MO-CNN)  and outperforms the classical ranking algorithms on the MORPH2 dataset. Similarly, a Ranking-CNN approach, that utilizes multiple deep binary classifiers, is proposed in \cite{Chen:CVPR17}. This method employs an ensemble of CNNs that are fused with aggregation. The inconsistency of the binary outputs is proven not to affect the overall performance as the output error is bounded by the maximum error of the binary classifiers \cite{Chen:CVPR17}. Therefore, as long as the maximum error is decreased, the inconsistent labels of the classifiers do not matter.

A different age ranker for separate age-groups is adopted in \cite{Liu:PR17} to learn a series of aging patterns. A deep CNN architecture is employed to minimize the distance of faces within the same age-group, while maximizing the distance between faces from different groups. The age prediction is then acquired using the OHRank algorithm for each age-group. To effectively maximize the distance between age groups, label-sensitive deep metric learning (LSDML) is introduced \cite{Liu:TIFS18}. This method optimizes the procedure by jointly learning a discriminative metric and mining hard negative pairs. In order to mitigate the effect of sparse and imbalanced datasets, the method is extended to multisource LSDML that maximizes the cross population correlation between different datasets.

A large number of deep methods are applied to apparent age estimation, mainly hosted by the Chalearn Looking at People (LAP) challenge  \cite{Escalera:ICCVw15}, \cite{Escalera:CVPRw16}. In particular, a deep network consisting of 10 convolutional layers, 5 pooling layer and 1 fully connected layer \cite{Chen:WACV16} is used in \cite{Ranjan:ICCVw15}. For the aging function, a test image is classified into 3 age groups and a 3-layer ANN regresses the apparent age. Each of the regressors is trained on a different dataset.

The highest positions in the LAP challenge are populated by variation of popular deep learning architectures. Such variations include ensembles of multiple networks and fusion of different models. The most widely adopted ones are the VGG-16 \cite{Simonyan:VGG16} and GoogLeNet \cite{Szegedy:GoogLeNet} Deep CNN, due to their success in the Imagenet object recognition challenge \cite{Russakovsky:Imagenet}. Modified versions of the original VGG-16 model are used for apparent age estimation in \cite{Rothe:IJCV16}, \cite{Kuang:ICCVw15}, \cite{Antipov:CVPRw16} and \cite{Uricar:CVPRw16}. These models are trained and finetuned on several different datasets to achieve significant generalization, while a variety of classification and regression techniques are employed on the deep representations.  Particularly, an SVM based classifier is used in \cite{Uricar:CVPRw16} and improves the results of the regression method used in \cite{Rothe:IJCV16}. Lastly, a fusion of GoogLeNet-based classifier and regressor is used in \cite{Liu:ICCVw15} and hierarchical SVM age grouping followed by SVR and RF regression on the deep features is employed in \cite{Zhu:ICCVw15}.

\subsection{Other representations}

Besides the above facial representation, several other approaches have been proposed. Such methodologies either utilize multiple descriptors, e.g., \cite{Bauckhage:ICPR10}, \cite{Chen:FG11}, \cite{Ersi:ICIP14}; that is, multiple features are fused to capture more discriminative information, or introduce novel extensions to these representations, e.g., \cite{Lu:TIP15}, \cite{Wang:TCyber15}. In particular, the Neighborhood Centroid Difference Vector (NCDV) from the LBP descriptor is used in \cite{Lu:TIP15} and the Cost-Sensitive Local Binary Features (CS-LBFL) are introduced. A multi-feature learning extension is also proposed and the final prediction is obtained using the OHRank algorithm.

Most descriptors in this section take into account age-sensitive information, e.g., wrinkles, smoothness, bags under eyes, facial hair, etc. Since different age groups share the same attributes, such information is not always accurate and can lead to misclassifications. Nevertheless the presence of these attributes is different between the age groups. In  \cite{Wang:TCyber15}, a Learning Using Privileged Information (LUPI) \cite{Vapnik:LUPI} approach is proposed, so that such inaccuracies are used to boost the model's generalization ability. That is, for each attribute, the age groups are ordered based on the presence of the specific attribute. The problem is solved using relative attribute SVM+ (raSVM+).

In order to obtain a discriminative mapping of the feature space, a number of the above methods have introduced  hierarchical features. The notion of hierarchical representations is not limited to deep learning methods, but is generally used in representations that accumulate information in multiple layers. A simple and efficient method to extract hierarchical features for age estimation is presented in  \cite{Kong:TIP15}. The method, namely MidFea, consists of k-means, convolution, max-pooling, vector quantization and random projection operations. The MidLevel features are then used as input to a Neuron Selectivity layer (NS) \cite{Bienenstock:NeuronSelect} to obtain the final representation.

All of the above representations originate from 2-Dimensional images; that is, they are either extracted from pixel intensities or facial landmarks. Nevertheless, other modalities, e.g., 3D \cite{Xia:VISAPP14}, video \cite{Dibekliouglu:TIP15}, context \cite{Gallagher:Groups}, contain age information and a number of methodologies are proposed to capture it. In particular, Gallagher et al. \cite{Gallagher:Groups} assign contextual features that capture local pairwise information and the global position of the person. A Gaussian Maximum Likelihood (GML) classifier is used to classify the faces into 7 age groups. Interestingly, the contextual features achieve an accuracy of more than double random chance, while gender is correctly estimated two-thirds of the time.

An interesting combination of different modalitites is that of appearance with facial dynamics. This is investigated in \cite{Dibekliouglu:TIP15}, where temporal face dynamics are  extracted from smile and disgust videos. The videos that are used include posed and spontaneous expression and were recorded under controlled conditions. The experiments in \cite{Dibekliouglu:TIP15} reveal that dynamic features are not as discriminative as appearance features for age estimation. Towards that end, several appearance estimators are used, i.e. dynamic, intensity based encoded features (IEF), gradient-based encoded features (GEF), BIF and LBP, while feature selection is performed using the Min-Redundancy Max-Relevance (mRMR) algorithm. The combination of the appearance and dynamic features achieves the best results, indicating that dynamics can help significantly towards age estimation.

\subsection{Evaluation protocols, metrics, and results for age estimation}

\textit{\textbf{Protocols: }} A number of different evaluation protocols have been used to evaluate Age Estimation models. Classical methods like k-fold cross-validation (c-v) and 80-20 split have been used in this context as well. Contrary to c-v, when applying an 80-20 split, a predefined 80\% of the data is used for training and 20\% is used for testing, without reshuffling the data. Besides these widely adopted protocols, specific protocols have been proposed for specific datasets.

One of the most widely used protocols is the Leave-One-Person-Out (LOPO) protocol, proposed for the FGNET dataset. According to this protocol, experimental evaluation is performed using images of previously unseen individuals. Therefore, training is done using all subjects in the database apart from the subject whose age we are estimating.

Other dataset-specific protocols include the Images of Groups (Groups) protocol and the Chalearn apparent age dataset protocol (chalearn). The data split for the Groups is a random selection of 3,500 training images and 1,050 testing images. The age group classification accuracy (acc.) is calculated both for exact match (AEM) as well as for allowing error of one category (AEO). On the other hand, the Chalearn dataset is split into 2,476 images for training, 1,136 images for validation and 1087 images for testing.

\textit{\textbf{Metrics: }}The large number of methods in age estimation has exposed the need for a common evaluation protocol. The most widely adopted measures are the Mean Absolute Error (MAE) and the Cumulative Score (CS). The MAE is defined as the average of the distance between the predicted age labels and the ground truth, 

\begin{equation}
MAE =\sum_{\j=1}^N
|\bar{y}-y|/N.
\end{equation} On the other hand, CS($\it{j}$) is the percentage of the samples, where the predicted age of model did not deviate from the ground truth more than $\it{j}$ years,\begin{equation}
CS(j) = N_{e\leq j} / N \times 100 \%.
\end{equation} Lastly, specifically for the apparent age estimation models \cite{Escalera:ICCVw15}, the adopted measure is the error calculated as follows: \begin{equation}
error =1 - e^{-\frac{(y-y^i)^2}{2\sigma^2}}.
\end{equation}

\textit{\textbf{Results: }}Table \ref{T:samll age} includes the methods that have reported the highest scores for each dataset in this subsection, using comparable and widely used protocols. A chronological overview of the most important methods and result on Age Estimation is tabulated in Table \ref{T:large age}. 

\begin{table}[htbp]
  \center
  \caption{Best reported results on Age Estimation for each dataset}
    \begin{tabular}{lllllllr}
    \toprule
    Dataset &method& metric & protocol& score \\
          \midrule
    FG-NET  & \cite{Liu:TIFS17} & MAE & LOPO  & 3.74 \\
    MORPH2 & \cite{Liu:TIFS17} & MAE & 80-20 & 2.89  \\
    Groups  &\cite{Li:CVPR12}& AEM/AEO acc. & groups & 48.5\% , 88\% \\
    Adience & \cite{Liu:TIFS17}   & acc. & 5-fold c-v & 60.2\% $\pm 5.3$\\
    AFAD & \cite{Niu:CVPR16} & MAE,CS($\leq{10}$) & 80-20 & 3.34, $\sim{95\%}$\\
    Chalearn & \cite{Antipov:CVPRw16}   & error & chalearn & 0.2411\\
    UvA-NEMO & \cite{Dibekliouglu:TIP15}   & MAE & 10-fold c-v & 4.33\\
    
    \bottomrule
    \end{tabular}%
  \label{T:samll age}%
\end{table}%

\subsection{Discussion on age estimation}

There is a number of factors affecting the accuracy of age estimation algorithms. The choice of facial representation plays a vital role in this, with the advantages of each approach having been  described above. Besides that, the types of variation within each dataset determines the success of the facial representation and the algorithm. Such variations include image quality, race, gender and facial expressions. An experimental study of age estimation under changes in image quality was conducted in \cite{Alnajar:ICIP15}. Image quality affects texture information that captures aging cues, like wrinkles. Therefore, the results confirmed the deterioration of the accuracy with declining image quality.

People of different gender do not follow the same aging patterns. A similar assumptions can be made for people of different race, eg. Caucasian and Asian. Several popular datasets, including MORPH2, contain faces from people of different gender and race. Such non linear variations affect the accuracy of the facial representations. The studies in \cite{Guo:CVPRw10} , \cite{Guo:CVPR14} and \cite{Bhattarai:ICASSP16} indicate this problem and propose transfer learning and domain adaptation solutions to deal with it. A similar problem arises when performing age estimation across different expressions. The quantitative study in \cite{Guo:CVPR12} indicates the significant influence of facial expressions on age estimation, while a cross-expression age estimation framework is proposed. 

In total, different datasets have different modes of variation, since they were collected under different circumstances. That is, image quality, illumination, race, gender and age among the data can vary significantly. Similar to the above, using the tools of transfer learning, Su et al.  represented different datasets as different domains and a cross-database age estimation framework is proposed in \cite{Su:ICASSP10}.

\subsection{Challenges in the wild}

\begin{figure*}[h!]
  \centering

    \includegraphics[width=\linewidth]{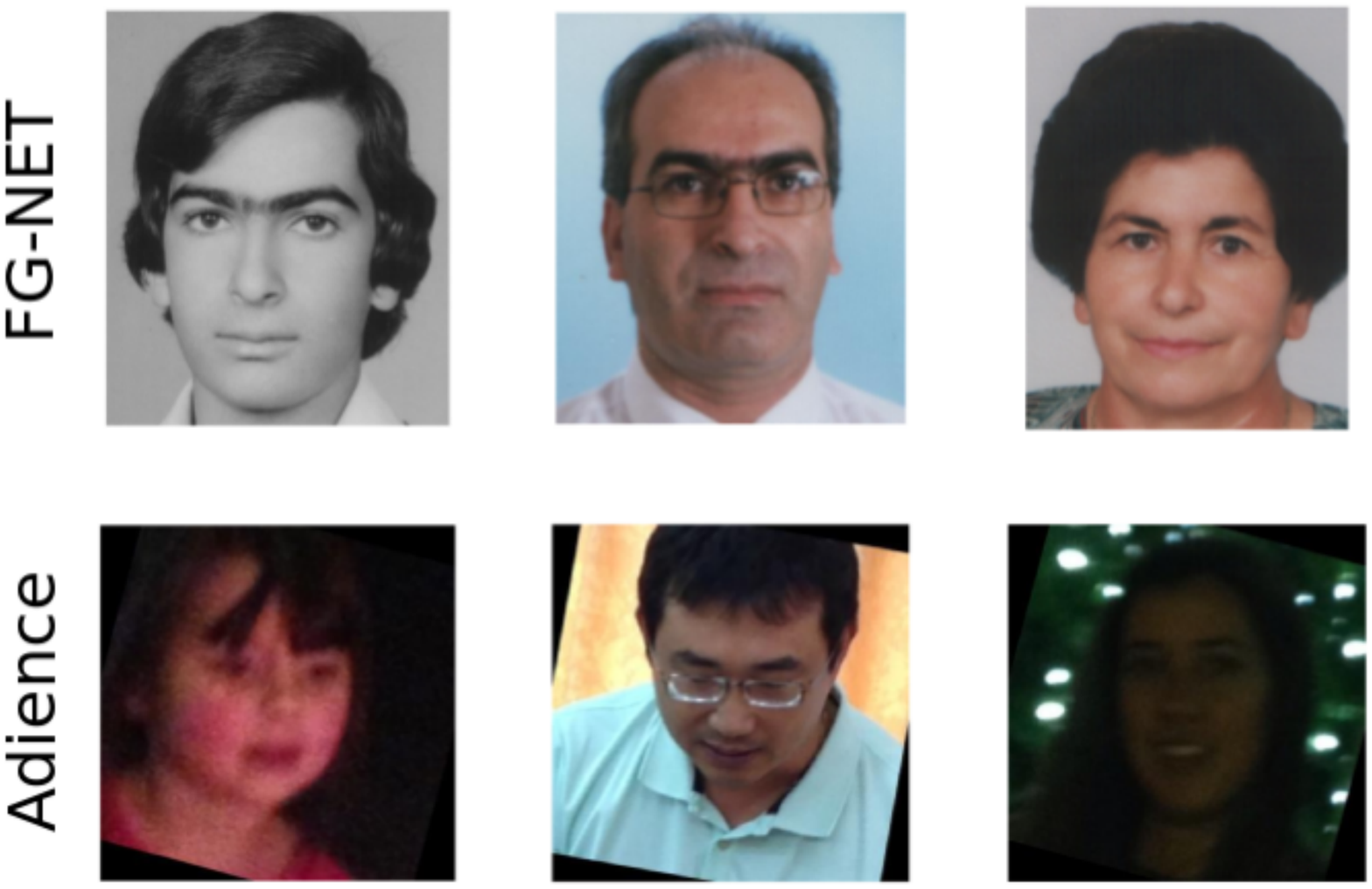}

  \caption{Samples from Adience and FG-NET datasets}
 
  \label{fig:differ_samples}
\end{figure*}

Most of the methods presented in this subsection are evaluated on in the wild datasets. Nevertheless, as mentioned in section \ref{S:data}, these datasets pose different levels of challenges. Table \ref{T:samll age} indicates that datasets like Adience and Groups are indeed more challenging, having great variation regarding face size, image quality, make-up and occlusions. Particularly, the median face from the Groups dataset has 18.5 pixels between the eye centers, while the Adience dataset demonstrates extreme variation in image quality (Figure \ref{fig:differ_samples}). On the other hand, FG-NET and MORPH2 contain family photos and mugshots respectively. While the image quality in the first one varies greatly, neither of them contain extreme poses. It is therefore expected, that the state-of-the-art results for these benchmarks are better compared to the Adience and Groups benchmarks. Lastly, the best score is achieved with the MORPH2 dataset, which can be justified by the fact that the faces in the mugshots are frontal.

\begin{table*}[]
\centering
%\begin{longtable*}{*6{p{2cm}}}
\begin{tabular}{ p{.08\textwidth}  p{.03\textwidth} p{.14\textwidth}  p{.20\textwidth}  p{.18\textwidth}  p{.08\textwidth}} 
\textbf{Year} & \textbf{Paper}                     & \textbf{Representation}               & \textbf{Method}                                       & \textbf{Dataset(Score)}                   & \textbf{Metric}     \\

\midrule
1994 & \cite{Kwon:CVPR94}        & ratios,wrinkles              & hand crafted rule                            & private(100\%)                   & accuracy   \\
\midrule
2002 & \cite{Lanitis:TPAMI02}    & AAM                          & regression                                   & private(1.88)                    & MAE        \\
\midrule
2005 & \cite{Zhou:ICCV05}        & Haar                         & regression, boosting                         & FG-NET(5.81)                     & MAE        \\
\midrule
2007 & \cite{Yang:Bio07}         & LBP                          & Rule-based, Adaboost              & FERET(7.88\%)                    & error rate \\
     &                           &                              &                                              & PIE(12.5\%)                      & error rate \\
%\midrule
2007 & \cite{Geng:TPAMI07}       & AAM                          & LDA, AGES                                    & FG-NET(6.22)                     & MAE        \\
     &                           &                              &                                              & MORPH(8.07)                      & MAE        \\
\midrule
2008 & \cite{Fu:TransMulti08}    & PI                           & CEA, multilinear regression                  & YGA-fem(8)                       & MAE        \\
     &                           &                              &                                              & YGA-male(7.8)                    & MAE        \\
%\midrule
2008 & \cite{Guo:TIP08}          & AAM                          & OLPP, LAR                                    & FG-NET(5.07)                     & MAE        \\
     &                           &                              &                                              & YGA-mal(5.30)                    & MAE        \\
     &                           &                              &                                              & YGA-fem(5.25)                    & MAE        \\
\midrule
2009 & \cite{Gallagher:Groups}   & contextual                   & GML                                          & Groups(32.9\%, 64.4\%)           & AEM, AEO   \\
%\midrule
2009 & \cite{Gao:ICB09}          & Gabor                        & Fuzzy LDA                                    & private(91\%)                    & accuracy   \\
%\midrule
2009 & \cite{Guo:CVPR09}         & BIF                          & PCA,SVM/SVR                                  & FG-NET(4.77)                     & MAE        \\
     &                           &                              &                                              & MORPH-fem(3.47)                  & MAE        \\
     &                           &                              &                                              & MORPH-mal(3.91)                  & MAE        \\
\midrule
2010 & \cite{Chang:ICPR10}       & AAM                          & Threshold ranker & FG-NET(4.67, $\sim{68\%}$)         & MAE,CS(5)  \\
     &                           &                              &                                              & MORPH(6.49, sim{50\%})           & MAE,CS(5)  \\
%\midrule
2010 & \cite{Zhang:CVPR10}       & AAM                          & Multi-task warped GP                         & FG-NET(4.14)                     & MAE        \\
     &                           &                              &                                              & MORPH(4.07)                      & MAE        \\
%\midrule
%2010 & \cite{Fang:ICPR10}        & LBP                          & OLPP, 3-layer NN                  & FG-NET(3.6387, \sim{68\%})       & MAE, CS(5) \\
\midrule
2011 & \cite{Chang:CVPR11}       & AAM                          & OHRank                                       & FG-NET(4.48, 74.7\%)             & MAE,CS(5)  \\
     &                           &                              &                                              & MORPH (5.88, 56.5\%)             & MAE,CS(5)  \\
%\midrule
2011 & \cite{Guo:CVPR11}         & BIF                          & Kernel PLS                                   & MORPH(4.18)                      & MAE        \\
%\midrule
2011 & \cite{Song:ICCV11}        & BIF                          & multiview regression                         & Custom(6.94 $\pm{0.07}$) & MAE        \\
%\midrule
2011 & \cite{Ni:TransMult11}     & BIF                          & RMIR                                         & FG-NET(8.37)                     & MAE        \\
     &                           &                              &                                              & MORPH(6.06)                      & MAE        \\
%\midrule
2011 & \cite{Fan:ICCV11}         & AAM                          & CPPN, kNN                                    & FG-NET(4.67)                           & MAE        \\
%\midrule
2011 & \cite{Yang:CVPR11}        & PI                           & 5-layer CPNN                                 & FG-NET(4.88)                     & MAE        \\
\midrule
2012 & \cite{Wu:TIFS12}          & geometric, HGPP              & SVM, PLS                  & FG-NET(6.76)                     & MAE        \\
%\midrule
2012 & \cite{Thukral:ICASSP12}   & geometric HGPP               & RVM                                          & FG-NET(6.2)                      & MAE        \\
%\midrule
2012 & \cite{Li:CVPR12}          & Gabor , contextual           & PLO, OHRank                                  & FG-NET(4.82)                     & MAE        \\
     &                           &                              &                                              & Groups(48.5\%, 88\%)             & AEM, AEO   \\
\midrule
2013 & \cite{Geng:TPAMI13}       & AAM , BIF                    & MFA, CPNN/IIS-LLD                           & FG-NET(4.76)                     & MAE        \\
     &                           &                              &                                              & MORPH(4.87 $\pm{0.31}$)            & MAE        \\
%\midrule
2013 & \cite{Chen:CVPR13}        & AAM                          & CA, SVR                                      & FG-NET(4.67, 74.5\%)             & MAE,CS(5)  \\
     &                           &                              &                                              & MORPH(5.88, 57.9\%)              & MAE,CS(5)  \\
%\midrule
2013 & \cite{Chao:PR13}          & AAM                          & LPP, lsRCA, kNN, SVR                         & FG-NET(4.38)                     & MAE        \\
%\midrule
2013 & \cite{Chen:TIFS13}        & AAM                          & ranking SVR                                  & FG-NET(4.56)                     & MAE        \\
     &                           &                              &                                              & MORPH1(5.41)                     & MAE        \\
     &                           &                              &                                              & MORPH2-Cau(4.42)                 & MAE        \\
\midrule
2014 & \cite{Eidinger:TIFS14}    & LBP, FPLBP                   & Dropout-SVM                                  & Adience(45.1\% $\pm{2.6}$)     & accuracy   \\
     &                           &                              &                                              & Groups(66.6\% $\pm{0.7}$)          & accuracy   \\

%\midrule
2014 & \cite{Yi:ACCV14}          & PI                           & CNN                                          & MORPH(3.64)                      & MAE        \\
\midrule
2015 & \cite{Kong:TIP15}         & PI                           & MidFea-NS, SVM                               & FG-NET(4.73)                     & MAE        \\
%\midrule
2015 & \cite{Dibekliouglu:TIP15} & dynamic, IEF, GEF, BIF, LBP & mRMR, SVM, SVR                               & UvA-NEMO(4.33 $\pm{4.06}$)         & MAE        \\

2015 & \cite{Chang:TIP15}        & ST                           & CSOHRank                                     & FG-NET(4.70, sim{75\%})          & MAE,CS(5)  \\
     &                           &                              &                                              & MORPH(3.82, sim{78\%})           & MAE,CS(5)  \\
%\midrule
2015 & \cite{Duong:CVPR15}      & PI, landmarks                & DBM                                          & FG-NET(5.28)                     & MAE        \\
%\midrule

2015 & \cite{Li:TransCyber15}   & BIF                          & PLO, OHRank                                  & FG-NET(1.306)                    & MAE        \\
     &                           &                              &                                              & Groups(0.864)          & MAE        \\
     &                           &                              &                                              & FACES-best(5.16)                 & MAE        \\
%\midrule
2015 & \cite{Han:TPAMI15}        & BIF                          & Adaboost, SVM,SVR                            & FG-NET(3.8 $\pm{4.2}$)             & MAE        \\
     &                           &                              &                                              & MORPH(3.5 $\pm{3.0}$)              & MAE        \\
     &                           &                              &                                              & PCSO(4.1 $\pm{3.3}$)               & MAE        \\

\midrule
2016 & \cite{Wang:TCyber15}      & DSIFT, LUPI                  & raSVM+, OHRank                               & FG-NET(4.07)                     & MAE        \\
     &                           &                              &                                              & MORPH(5.05$\pm{0.11}$)             & MAE        \\
%\midrule
2016 & \cite{Niu:CVPR16}         & PI                           & MOCNN                                        & MORPH(3.27)                      & MAE        \\
     &                           &                              &                                              & AFAD(3.34)                       & MAE        \\
%\midrule
2016 & \cite{Rothe:IJCV16}       & PI                           & DEX                                          & Chalearn(3.221, 0.2649)          & MAE, error \\
%\midrule
2016 & \cite{Antipov:CVPRw16}    & PI                           & ensemble DEX                                 & Chalearn(0.2411)                 & error      \\
\midrule
2017 & \cite{Chen:CVPR17}    & PI                           & Ranking-CNN                                 & MORPH(2.96, $\sim{85\%}$)                 & MAE,CS(5)      \\
2017 & \cite{Liu:TIFS18}    & PI                           & M-LSDML                                 & MORPH(2.89)                 & MAE      \\
&                           &                              &                                              & Adience(60.2$\pm{5.3}$)             & accuracy        \\
&                           &                              &                                              & FG-NET(3.74)             & MAE        \\
&                           &                              &                                              & FACES(3.11 - 5.01)             & MAE        \\
&                           &                              &                                              & Chalearn(0.315)             & error        \\
2017 & \cite{Hu:TIP17}    & PI                           & AEAD                                 & MORPH(2.78)                 & MAE      \\
&                           &                              &                                              & FG-NET(2.8)             & MAE        \\

\end{tabular}
\caption{Overview of Age Estimation methods}
\label{T:large age}
\end{table*}

\section{Age progression} \label{S:age prog}

Face synthesis of an individual at different age groups is a task that has been studied for several years across computer graphics, anthropometry and computer vision. Although earlier methods used face models introduced in computer graphics, recent work has gradually transitioned to computer vision. One of the seminal works is by Burt et al. \cite{Burt:Biological95}, which simulates the aging process by employing both colour and shape information. In this approach, the synthesized image is produced by adding the difference between the average faces (i.e. the prototypes) of each age group to the test image. Other early works include anthropometric growth \cite{Ramanathan:CVPR06}, \cite{Ramanathan:FG08}, modeling of wrinkles \cite{Wu:94}, \cite{Wu:95}, \cite{Wu:99}, \cite{Tiddeman:CGA01}, aging functions \cite{Lanitis:TPAMI02prog} and caricaturing 3D face models \cite{OToole:Perception97}, \cite{OToole:IMAVIS99}. Notably, not all of the aforementioned earlier methods produce photorealistic results. More recently proposed methodologies in age synthesis are presented in the following paragraphs.

\subsection{Subspace Learning based Methods}

The tools provided by subspace and factor analysis have been used to model the transformations caused by aging. A prototype based approach that compensates for illumination variation in the prototypes is introduced in \cite{Ira:CVPR14}. In this Illumination Invariant Age Progression (IAAP) method, the prototype for each age cluster is obtained as a rank-4 approximation using Singular Value Decomposition (SVD) and  are relighted according to the input face. Then the input face is rendered to a different age by applying the difference in flow and texture between the source and target prototypes. The results are evaluated in a large scale user study on Mechanical Turk, the result of which indicates the inability of the workers to recognize a face across large age differences.

In prototype-based approaches (e.g., \cite{Burt:Biological95}, \cite{Ira:CVPR14}), the texture transformation is modelled as the difference between the prototypes of the source and target age groups. That is, the texture changes are the same for different inputs, as long as the source and target age is the same. Instead of a specific prototype, a dictionary of aging patterns for each age group is learned in \cite{Shu:ICCV15}. To capture the different factors of variation, an input face is decomposed into an aging and a personalized layer. The changes in the aging layer are modelled using the dictionaries of the target and the source age groups. A more general approach that is robust to different kinds of variation, namely Robust Age Progression (RAP), is introduced in \cite{Sagonas:ICPR16}. Each image of a specified gender is expressed as a superposition of the age component and the common component. The common component captures facial variations such as identity, shape, pose, occlusions, illumination and expressions, while the age information is captured by the age component. Thus, by computing an orthonormal bases of the age and common components via SVD, an image can be progressed to another age group as a linear combinations of these bases at the \textit{k}th age group.

In a similar manner, Hidden Factor Analysis (HFA) is used in \cite{Yang:TIP16} to decompose the facial input. Thus, the face is decomposed into a linear combination of the mean face and the identity, age and noise factors. The age component at a different age is then sparsely represented by a dictionary of age components of the same age group. The shape is progressed by applying the difference between the mean shapes of the target and source age clusters. In order to disentangle the common and individual components, a robust extension to Joint and Individual Variation Explained \cite{Lock:JIVE} is used in \cite{Sagonas:CVPR17}. Thus, by modifying these components, an input face can be reconstructed at a different age group. The method is evaluated on the FG-NET and AgeDB datasets, while experiments are conducted on facial expression synthesis as well.

To deal with the different modes of variation, instead of a matrix, a tensor is used to represent the data in \cite{Wang:TSMC12}. The different modalities of the data, namely the pixels intensities, identities and ages of the face images, are included along the the dimensions of the tensor. This means that along each axis of the 3-dimensional structure, the variation of only one modality changes. Based on the fact that low frequency image components preserve the face identity, super-resolution in tensor space is used to map the texture of a downsampled input image to a different age.

\subsection{Sequence modeling}

Since the effects of aging are temporally correlated, i.e. the transformation is smooth and continuous, it is intuitive to model it as a sequence across the different age groups. In order to describe the evolution of the facial representation through the age groups, different sequence modeling methods have been proposed, including Markov Process \cite{Suo:TPAMI10}, \cite{Suo:TPAMI12} and Recurrent Neural Networks (RNN) \cite{Wang:CVPR16}. In particular, the transformation of different facial parts is modelled separately in \cite{Suo:TPAMI12}. Instead of using AAM, a region based AAM model (RB-AAM) is adopted. The relationships between the different sub-regions are modelled based on the physical structure of the face. To approximate the short-term aging of each sub-region, an aging function approach is used. The aging model over a long period is formulated as a Markov Process by concatenating the short period aging functions under smoothness and consistency constraints.  On the other hand, Wang et al. \cite{Wang:CVPR16} use RNN \cite{Graves:LSTM} to model age progression between neighbouring age groups. The faces of the neighbouring groups are normalized jointly. An RNN is adopted to  perform aging of a face in the shared eigenface space \cite{Turk:Eigenfaces} of two neighbouring age groups. The output of the RNN is reconstructed, projected on the shared eigenface space of the next group and used as input for the next RNN.

\subsection{3-Dimensional Representations}

The aforementioned methods come with certain limitations since they are based on 2-dimensional representations. Contrary, 3D facial representations capture both shape and texture information and can potentially obtain pose invariance. A 3-dimensional model is used to build an age progression system for children faces \cite{Shen:ISM11}. Different facial components, such as mouth, eyes, nose and face shape, are extracted and progressed individually. The basic assumption is that if two children look similar at a young age, they will continue to look similar when they grow older. Therefore, each component is compared to a database of facial components and is progressed according to the corresponding aging pattern. The selected components are then merged to synthesize the progressed image. The experimental results on selected images of the Jackson family indicate the possibility of synthesizing age progressed faces of children given the faces of their relatives.

A different approach that incorporates the whole face in a 3D aging model is proposed in \cite{Park:TPAMI10}. The aging model is approximated as a weighted average of all the texture and shape aging patterns in the training set. The model is evaluated using a commercial face recognition system as described in Section 6. Similarly, texture and shape are modelled separately in \cite{Maronidis:VISAPP13}. The face shape of each age group is modelled using age specific 3D Point Distribution Model (3DPDM) \cite{Cootes:CVIP95}, while the texture is modelled using recursive PCA.

\subsection{Deep learning-based representations}

Synthesizing a photorealistic face image with arbitrary modes of variation is a challenging task. Deep learning methods are able to incorporate knowledge from multiple datasets and are therefore suited to deal with this complicated problem. In particular, Generative Adversarial Networks (GANs)\cite{Goodfellow:NIPS14} have proven capable of producing realistic images and have been successfully applied to face synthesis, e.g., \cite{Liu:NIPS16}, \cite{Zhao:EBGAN}. The original model consists of two networks, a Generator and a Discriminator, that are trained simultaneously. The Generator tries to model the data distribution and synthesizes images which the Discriminator classifies as 'real' or 'fake', that is whether they come from the actual data distribution. The image generation is conditioned on random input noise \textit{z}, which follows a predefined distribution. The optimization procedure can be considered as a minmax game between the two models, as the parameters of each network are optimized alternatively. In some variations of the model, particularly in the Conditional GAN (CGAN) \cite{Mirza:CGAN}, the generated images are conditioned on a specific input, rather than random noise.

A Conditional Adversarial Autoencoder (CAAE) for age progression is introduced in \cite{Zhang:CVPR17}. Instead of randomly sampling \textit{z}, it is obtained from an autoencoder, so that incorporates the personality of the face. The model includes two discriminators, one to enforce \textit{p(z)} to be uniform; that is, to force \textit{z} to evenly populate the latent space, and one to force the generator to produce realistic images. The generation of the image is conditioned on both the personality of the input, as well as the desired age group. The method is evaluated based on survey results, that indicate the ground truth. A conditional GAN (Age-cGAN) is also employed in \cite{Antipov:ICIP17} to perform face aging.  The model obtains the embeddings of the input and ouput of the GAN from a face recognition neural network. In order to preserve the identity of the input face, the Euclidean distance between the embeddings is minimized.

\subsection{Evaluation protocols, metrics, and results for age progression} \label{subS:eval age prog}

\textit{\textbf{Protocols:}} There is a wide variety of different protocols in the literature regarding the evaluation of age progression models, since not all methods focus on the same age groups, e.g., some focus on children faces, while others focus on adult faces. That being said, all the methods in this subsection use the FG-NET dataset to test their methods qualitatively or quantitatively.

\textit{\textbf{Metrics:}} The evaluation of age progression methods is nontrivial, since the task itself is ill-posed and the actual aging process is highly uncertain. A comparative study of different age progression evaluation techniques is presented in \cite{Lanitis:EURASIP08}, while a framework for evaluation is described in \cite{Lanitis:FG15}. A short description of the most common evaluation methods is presented as follows:

\textit{a. Human-based evaluation:}
Directly comparing the synthesized images with the ground truth is not an effective measure of accuracy, as the images were taken under different sources of variation. A way to evaluate the results of age progression algorithms is by conducting a study with human users. The results can then be evaluated based on the users' answers to questions comparing the synthesized image and the ground truth at the target age.

\textit{b. Age estimation accuracy:}
Age progression models are also evaluated based on their ability to generate faces with characteristics of the target age group. The accuracy of the synthesis can be measured in an automated manner, using an age estimator and the corresponding metrics as described in Section 4. 

\textit{c. Preservation of face identity:}
An age progression method is evaluated based on whether the synthesized images can still be recognized as images of the same person. To determine that, face recognition and verification systems have been widely used. In the context of age progression evaluation, common metrics include classification accuracy (rank n), the ROC curve and the area under it (AUC) as well as Equal Error Rate (EER). A more detailed description of Age-Invariant Face Recognition (AIFR) and Cross-Age Face Verification (CAFV) methods and the corresponding metrics is included in Section 6.

\textit{\textbf{Results:}} While the methods in this subsection have been evaluated on different datasets, we will only use the common results on FG-NET for comparison. Based on their quantitative results on FG-NET, the most significant results are presented in Table 5. The results for preservation of identity are reported before (score-b) age progression as well as after (score-a) wherever available.

\begin{table}[htbp]
  \center
  \caption{Best reported results on Age Progression based on preservation of identity on FG-NET}
    \begin{tabular}{lllllllr}
    \toprule
    method& test & metric &score-b& score-a \\
          \midrule
    \cite{Shu:ICCV15} & CAFV&EER & 14.89\%  & 8.53\% \\
    \cite{Sagonas:ICPR16} &CAFV& acc., AUC & N/A & 0.709, 0.806   \\
    \cite{Wang:TSMC12}&AIFR&  accuracy & 0.55 & 0.7 \\
    \cite{Wang:CVPR16} & CAFV & EER & $\sim{15\%}$& $\sim{10\%}$\\

    \bottomrule
    \end{tabular}%
  \label{tab:addlabel}%
\end{table}%

Results on other databases are scarce. Preservation of identity results on AgeDB, Morph, CACD and Browns are tabulated in Table \ref{T:other progression}. The results on AgeDB are reported for age difference of 10 years. A chronological overview of the methods in this subsection is tabulated in Table \ref{T:large prog}.

\begin{table}[htbp]
  \center
  \caption{Best reported results on Age Progression based on preservation of identity on other datasets, score-B and score-A correspond to face verification score before and after age progression}
    \begin{tabular}{lllllllr}
    \toprule
    dataset& meth.& test & metric &score-B& score-A \\
          \midrule
    AgeDB&\cite{Sagonas:CVPR17} & CAFV& acc.,AUC & 0.591,0.624   & 0.621,0.654  \\
    MORPH&\cite{Park:TPAMI10} & CAFV& acc.(r=5) & 68\%  & 73\% \\
    CACD&\cite{Sagonas:ICPR16} &CAFV& acc.,AUC & N/A & 0.735, 0.798   \\
    BROWNS&\cite{Park:TPAMI10}&CAFV& acc.(r=5) & 45\% & 60\% \\

    \bottomrule
    \end{tabular}%
  \label{T:other progression}%
\end{table}%

\subsection{Discussion on age progression methods}

Since age progression is a challenging task, age clusters are labelled with whole age-groups and not exact ages. This formulation also helps with the highly incomplete datasets that are available. This issue is more dominant in faces of children, where images are very hard to collect and old family photos are often of bad quality.

\subsection{Challenges in the wild}

The number of widely adopted benchmarks for age progression is limited to 5 datasets, all of which are captured in the wild. The use of the CAFV protocol does not allow for quantitative comparison between the different datasets, since different face verification algorithms are used. Therefore, a comparison can only take place using qualitative standards that are hard to evaluate. The challenging nature of the task has led research to focus mainly on datasets with small variation in pose and image quality. The qualitative differences between the results is largely attributed to the methods and so, conclusions can not be drawn regarding the 'wildness' of the datasets for the task of age progression.

\begin{table*}[]
\centering
\begin{tabular}{lllllll}
\textbf{Year} & \textbf{Paper}                    & \textbf{Repr.} & \textbf{Method}                                & \textbf{Dataset(score-B - score-A)}     & \textbf{Test}         & \textbf{Metric}        \\
\midrule
1995 & \cite{Burt:Biological95} & 2D    & prototype based                       & private                         &     N/A         &         N/A      \\
\midrule
1999 & \cite{OToole:IMAVIS99}   & 3D    & PCA                                   & private                         &      N/A        &      N/A         \\
\midrule
2001 & \cite{Tiddeman:CGA01}    & 2D    & wavelet, prototype based              & private                         &      N/A        &     N/A          \\
\midrule
2006 & \cite{Ramanathan:CVPR06} & 2D    & craniofacial growth                   & Custom(28\% - 37\%)            & Recogntion   & acc.(rank=1)  \\
\midrule
2008 & \cite{Ramanathan:FG08}   & 2D    & craniofacial, wrinkle analysis & Custom(38\% - 51\%)            & Recogntion   & acc.(rank=5)  \\
\midrule
2010 & \cite{Suo:TPAMI10}       & 2D    & And-Or Graph, Markov Chain                        & LHI                             &       N/A       &     N/A          \\
     &                          &       &                                      & MORPH                           &      N/A        &    N/A           \\
2010 & \cite{Park:TPAMI10}      & 3D    & weighted average                      & FG-NET(42\% - 55\%)             & Verification & acc.(rank=5)  \\
     &                          &       &                                       & MORPH(68\% - 73\%)              & Verification & acc.(rank=5)  \\
     &                          &       &                                      & BROWNS(45\% - 60\%)             & Verification & acc.(rank=5)  \\
\midrule
2011 & \cite{Shen:ISM11}, \cite{Shen:JISE14}        & 3D    & metric learning                       &      FG-NET                           &       N/A       &     N/A          \\
 &       &     &                      &      Jackson Family                           &       N/A       &     N/A          \\
\midrule
2012 & \cite{Wang:TSMC12}       & 2D    & tensor space analysis                  & FG-NET(55\% - 70\%)             & Recogntion   & Accuracy      \\
2012 & \cite{Suo:TPAMI12}       & 2D    &  RB-AAM, Markov Chain                        & Custom             &      N/A        &       N/A        \\
\midrule
2014 & \cite{Ira:CVPR14}        & 2D    & IAAP                       & FG-NET                          &         N/A     &        N/A       \\
%2014 & \cite{Shen:JISE14}       & 3D    & metric learning                       & FG-NET                          &      N/A        &      N/A         \\
%     &                          &       &                                      & Jackson Family                  &        N/A      &       N/A        \\
\midrule
2015 & \cite{Shu:ICCV15}       & 2D    & dictonary learning                    & FG-NET(14.89\% - 8.53\%)        & Verification & EER           \\
\midrule
2016 & \cite{Yang:TIP16}    & 2D    & HFA                  & FG-NET($\sim{52\%}$  $-$   $\sim{54\%}$)      & Recognition & acc.(rank=1) \\
2016 & \cite{Sagonas:ICPR16}    & 2D    & RAP                    & FG-NET(N/A - 0.709)      & Verification & Accuracy \\
 &        &     &                    & FG-NET(N/A-0,806)        & Verification &  AUC           \\
     &                          &       &                                       & CACD(N/A - 0.735)        & Verification & Accuracy \\
      &        &     &                    & CACD(N/A-0,798)        & Verification &  AUC           \\
2016 & \cite{Wang:CVPR16}       & 2D    & RFA                                 & FG-NET($\sim{15\%}$ -  $\sim{9\%}$) & Verification & EER       \\
\midrule
2017 & \cite{Zhang:CVPR17}       & 2D    & CAAE                    & Custom        & N/A & N/A           \\
2017 & \cite{Sagonas:CVPR17}       & 2D    & RJIVE                    & AgeDB(0.591- 0.621)        & Verification & Accuracy           \\
 &        &     &                    & AgeDB(0.624-0,654)        & Verification &  AUC           \\
 2017 & \cite{Antipov:ICIP17}       & 2D    & Age-cGAN                    & IMDB-WIKI-cleaned( - 82.9\%)        & Recognition & accuracy           \\
\end{tabular}
\caption{Overview of Age Progression Methods, score-B and score-A correspond to face verification score before and after age progression}
\label{T:large prog}
\end{table*}

\section{Age-invariant facial characterization} \label{S:age inv}

Age-invariant facial characterization involves two basic tasks: age-invariant face recognition and cross-age face verification. The goal of the methods for age-invariant face recognition is to build a model that is able to recognize the identity of a face across different ages from a database of faces. On the other hand, cross-age face verification aims to determine whether two age-separated images are from the same person. These tasks can be approached either with generative or discriminative methods. In generative approaches, an input face is transformed to the target age before performing face recognition, according to the methods described in the section \ref{S:age prog}. On the other hand, in discriminative approaches \cite{Ling:ICCV07}, \cite{Biswas:BTAS08}, age-invariant representations are extracted and a classification problem is solved. The generative methods are described in the previous section, while the most recent discriminative methods are presented in the following.

\subsection{Age-invariant descriptors}

The methods in this subsection focus on obtaining features that are invariant to the aging process of the face. The local descriptors that are widely adopted in face recognition, e.g., LBPs, have been used in an age-invariant context as well. An experimental evaluation of local descriptors in age-invariant face recognition is presented in \cite{Bereta:PR13}. Nevertheless, these classical descriptors cannot be used as stand-alone facial representations, as they do not always capture the age-invariant information. Therefore, the following methods focus on introducing new facial representations that are robust to age changes. The face representation used in \cite{Ling:TIFS10} is a hierarchical combination of Gradient Orientations (GO) \cite{Chen:CVPR00} of each color channel of the image at different scales, called Gradient Orientation Pyramid (GOP). The hierarchical information captured by this descriptor is classified using SVM and improves the verification results on adults. On the other hand, higher order information  does not improve the accuracy significantly for age changes in teenagers.
 
Using high dimensional LBPs, an age-invariant representation for cross-age face verification is described in \cite{Chen:ECCV14}, \cite{Chen:TransMult15}. After extracting the descriptors, a reference representation of every person is obtained at each age. The features are then encoded into this reference space and max pooling is used to normalize the representations of the same person at different ages. The resulting features are age-invariant, since they have a high response at the reference person at any year. The face pairs are then classified according to their cosine similarity. Lastly, a novel encoder that makes use of binary patterns, similarly to LBPs, is introduced in \cite{Gong:CVPR15}. Unlike other encoders, this descriptor converts binary patterns to evenly distributed codes. The final representation is obtained by maximizing the entropy of the descriptor. Identity matching is performed by decomposing the representation using Identity Factor Analysis (IFA) and classifying the cosine similarity of the inputs. A similar feature extractor, called Local Pattern Selection (LPS), is used at multiple scales in \cite{Li:TIP16}. The multiple scaling and dense sampling of the method result in a high-dimensional representation, which is refined using bagging \cite{Breiman:Bagging} multiple classifiers.

\subsection{Age-invariant subspace learning-based methods}

Similar to the previous tasks, instead of focusing on age-invariant local descriptors, an age-invariant subspace can be learned from simple representations. To build this subspace, several subspace analysis methods have been employed. In particular, Multi-Feature Discriminant Analysis (MFDA) is employed in \cite{Li:TIFS11}. The subspace is obtained from Scale Invariant Feature Transform (SIFT) \cite{Lowe:SIFT} and Multi-Scale LBPs features. The classification problem is solved in the subspace using bagging. The LBP feature space is also employed in \cite{Bouchaffra:TNN15} to obtain a subspace that captures geometric features of the data, such as shape. Nonlinear Topological Component Analysis (NTCA) is introduced to obtain the low-dimensional age-invariant subspace.

Another subspace learning method that has been used in other age modeling tasks, e.g., age progression \cite{Yang:TIP16}, is the HFA. Hidden Factor Analysis decomposes facial features into a linear combination of the age component, the identity component and the noise term. Contrary to age progression, for the task of face recognition, the identity component is taken into account. The feature space is obtained from Histograms of Oriented  Gradients (HOGs) \cite{Dalal:HOG}. The final output is obtained based on the cosine similarity of the identity components.

Lastly, a different method that coordinates cross-age face verification and cross-face age verification is introduced in \cite{Du:CVPR15}.  Motivated by the fact that, the first task needs age-invariant features while the latter needs age-sensitive features, the method coordinates the two in a multi-task learning manner. To achieve this, both tasks share the same feature pool and feature interaction is encouraged via an orthogonal regularization. That is, the final features are selected so that the age-sensitive ones are avoided.

\subsection{Deep learning-based methods}

Deep Learning models have not found wide use for age-invariant facial characterization, possibly due to the lack of large aging datasets. Nevertheless, a latent factor guided convolutional network is presented in \cite{Wen:CVPR16}. The fully connected layer of the CNN is designed using Latent Identity Analysis, separating the identity components from the rest of the convolutional features. The experiments performed indicate that simply finetuning a deep CNN on an aging dataset improves the accuracy significantly. This reveals that a baseline CNN needs further processing in order to efficiently learn age-invariant features.

\subsection{Evaluation protocols, metrics, and results for age-invariant facial characterization}

\textit{\textbf{Protocols:}} The most widely used datasets in this subsection are the FG-NET, MORPH and CACD datasets, while the protocols involve some of the aforementioned ones like k-fold cross-validation and LOPO. Particularly for the MORPH Album 2 dataset, a common protocol includes splitting the dataset into a training and a test subset, each containing images from 10,000 subjects (10k-10k). Lastly, it should be noted that since the CACD dataset consists of some noisy labels as well as duplicates, a verification subset called CACD-VS is introduced in \cite{Chen:TransMult15} and used for testing.

\textit{\textbf{Metrics :}}
Different methods have been used to evaluate age-invariant facial recognition and cross-age face verification. Some methods \cite{Du:CVPR15}, \cite{Ling:TIFS10} use Equal Error Rate (EER) to evaluate the performance of the verification. This metric describes the rate at which both accept and reject errors agree. Nevertheless, the most popular evaluation metric is verification accuracy or recognition accuracy. In most cases, rank-1 or rank-n accuracy is used, where the rank indicates the number of gallery images that have to be inspected in order to achieve this performance.

\textit{\textbf{Results:}} 
The most accurate method among the ones included in this subsection is the deep learning model in \cite{Wen:CVPR16}, which outperforms the rest in all datasets. The results are tabulated as follows, along with the second-best performance in each dataset. A chronological overview of the methods in this subsection is tabulated in Table \ref{T:large inv}.

\begin{table}[htbp]
  \center
  \caption{Best reported results on age-invariant facial characterization}
    \begin{tabular}{lllllllr}
    \toprule
    dataset& method& metric & protocol &score \\
          \midrule
    FG-NET&\cite{Wen:CVPR16} & rank-1 acc.& LOPO& 88.1\%  \\
    FG-NET&\cite{Gong:CVPR15} & rank-1 acc.& LOPO & 76.2\%  \\
    MORPH&\cite{Wen:CVPR16} & rank-1 acc.& 10k-10k & 97.51\% \\
    MORPH&\cite{Li:TIP16} & rank-1 acc. & 10k-10k & 94.87\% \\
    CACD-VS&\cite{Wen:CVPR16} &rank-1 acc.& 10-fold c-v & 98.5\%  \\
    CACD-VS&\cite{Chen:TransMult15} &rank-1 acc.& 10-fold c-v & 87.6\%  \\

    \bottomrule
    \end{tabular}%
  \label{T:small_inv}%
\end{table}%

\subsection{Discussion on age-invariant characterization methods}

Similar to classical Face Verification and Recognition methods, pose, illumination, occlusions and expressions are significant variation inducing factors. Therefore, the aforementioned methods should be able to obtain invariant representations, not only to aging, but also these kind of variation. The issue of inadequate aging datasets, that was mentioned in the previous subsection, stands for this task as well. This is important, especially for images of children, where experiments \cite{Ling:TIFS10} have shown that verification is much harder compared to adults. Lastly, face recognition and verification depend prominently on the age difference between the probe image and the gallery image \cite{Guo:ICPR10}, where the accuracy of the algorithm usually decreases with larger age gaps.

\subsection{Challenges in the wild}

Similar to age progression, age-invariant facial characterization methods have been evaluated only on three in the wild datasets. Since all three benchmarks follow similar protocols, they can be compared based on the the state-of-the-art results. The results on Table \ref{T:small_inv} validate the non-challenging nature of the MORPH2 benchmark, particularly in comparison to the pose and image quality variation in FG-NET.

\begin{table*}[]
\centering
\begin{tabular}{llllll}
\textbf{Year} & \textbf{Paper}                                       & \textbf{Representation}               & \textbf{Method}        & \textbf{Dataset(score)}           & \textbf{Metric}       \\
\midrule
2007 & \cite{Ling:ICCV07}                          & GOP                 & SVM           & Private I(5.1\%)         & EER          \\
     &                                             &                     &               & Private II(10.8\%)       & EER          \\
\midrule
2008 & \cite{Biswas:BTAS08}                        & SIFT                & Feature Drift & FG-NET(N/A)              & ROC          \\
\midrule
2010 & \cite{Ling:TIFS10}                          & GOP                 & SVM           & FG-NET($\geq{18}$)(24.1\%) & EER          \\
     &                                             &                     &               & FG-NET(8-18)(30.5\%)     & EER          \\
     &                                             &                     &               & FG-NET($\leq{8}$)(38.6\%)  & EER          \\
     &                                             &                     &               & Private I(5.1\%)         & EER          \\
     &                                             &                     &               & Private II(10.8\%)       & EER          \\
\midrule
2011 & \cite{Li:TIFS11}                            & SIFT, MLBP          & MFDA          & FG-NET(47.5\%)           & acc.(rank=1) \\
     &                                             &                     &               & MORPH(83.9\%)            & acc.(rank=1) \\
\midrule
2013 & \cite{Gong:ICCV13}                          & HOG                 & PCA, LDA, IFA & FG-NET(69\%)             & acc.(rank=1) \\
     &                                             &                     &               & MORPH(91.14\%)           & acc.(rank=1) \\
\midrule
2014 & \cite{Chen:ECCV14}, \cite{Chen:TransMult15} & CARC                & PCA,LDA, SVM  & MORPH(92.8\%)            & acc.(rank=1) \\
     &                                             &                     &               & CACD(87.6\%)             & acc.(rank=1) \\
\midrule
2015 & \cite{Gong:CVPR15}                          & MEFD, MLBP, SIFT    & PCA, LDA, IFA & FG-NET(76.2\%)           & acc.(rank=1) \\
     &                                             &                     &               & MORPH(94.59\%)           & acc.(rank=1) \\
2015 & \cite{Bouchaffra:TNN15}                     & LBP                 & NTCA          & FG-NET(48.96\%)          & acc.(rank=1) \\
     &                                             &                     &               & MORPH(83.8\%)            & acc.(rank=1) \\
2015 & \cite{Du:CVPR15}                            & SIFT, LBP, GOP, BIF & AGCD          & FG-NET(19.4\%)           & EER          \\
     &                                             &                     &               & MORPH(5.5\%)             & EER          \\
\midrule
2016 & \cite{Li:TIP16}                             & LPS                 & LFDA, HFA     & MORPH(94.87\%)           & acc.(rank=1) \\
2016 & \cite{Wen:CVPR16}                           & PI                  & LF-CNN        & FG-NET(88.1\%)           & acc.(rank=1) \\
     &                                             &                     &               & MORPH(97.51\%)           & acc.(rank=1) \\
     &                                             &                     &               & CACD(98.5\%)             & acc.(rank=1) \\
\end{tabular}
\caption{Overview of age-invariant facial characterization methods}
\label{T:large inv}
\end{table*}

\section{Kinship Verification}

Kinship Verification has recently gained interest as a task in  machine learning and computer vision community. The task refers to bi-subject verification; that is, classifying a pair of input images as kin or non-kin. A smaller number of methods have explored tri-subject kinship verification \cite{Qin:TransMulti15}, which compares a couple (Mother and Father) against a child, or whole family classification \cite{Fang:ICIP13}, as well as kinship recognition \cite{Guo:ICPR14}. The seminal work on kinship verification is by Fang et al. \cite{Fang:ICIP10}. The first kinship annotated dataset is introduced and facial parts (FP), such as mouth hair and nose, are extracted using a pictorial structure model \cite{Felzenszwalb:IJCV05}. That is, the facial structures are represented as parts in a deformable configuration. The representation also included color information, facial distances (FD) and gradient histograms (GH). The difference between the feature vectors of the query couple are classified using methods like k-NN and SVM. The experiments performed in the paper indicate that the most discriminant feature was the color of the eyes. Moreover, the highest accuracy was achieved for Father-Son image-pairs, while the algorithm outperformed the human workers. Such results suggest that kinship verification of other people is a difficult task for humans. Similar to the above, this section is organized based on the facial representation used for the task.

\subsection{Invariant descriptors to genetic variations}

Intuitively, people perceive kinship based on local facial attributes. These attributes often correspond to facial parts, like nose, eyes and mouth, the heredity of which indicates kin-related people. Different descriptors have been applied to capture facial part information for kinship verification. The eyes, nose and mouth are explored in \cite{Guo:TIM12}, where the DAISY \cite{Tola:TPAMI10} descriptor is used to represent each part. A similarity score is then computed for each part of the two faces to determine the familial traits. Similarly, 12 facial parts are explored in \cite{Fang:ICIP13} using the dense SIFT (dSIFT) descriptor. In this approach, each part is reconstructed from a dictionary of facial parts. The dictionary contains facial parts from the family of the query face, as well as non-related faces. In order to determine the family of a test face, sparsity is enforced on the reconstruction coefficients, based on the assumption that the query face only inherits features from its relatives. The final family prediction is obtained based on the three most dominant facial attributes.

A set of overlapping facial patches is used for feature extraction in \cite{Xia:ICPR12}. The face is segmented in 5 layers from coarse to fine; that is, the first layer contains the whole face and the other layers contain increasingly smaller parts. Binary attributes that indicate the presence or absence of a trait (e.g., mustache), as well as relative attributes (e.g., bigger nose) are taken into consideration. The LIFT algorithm \cite{Zhang:TPAMI15} is employed to extract the binary attributes, while a ranking function is learned for the relative ones. Finally, the output is obtained from an SVM classifier.

Instead of modeling different facial parts, several methods employ texture descriptors that are widely adopted in face verification. In particular, Local Phase Quantization (LPQ) \cite{Ojansivu:ICISP08}, Three and Four-Patch LBPs and Weber Local Descriptors (WLD) \cite{Chen:TPAMI10} are employed to represent the face in \cite{Bottinok:FG15}. The fusion of multiple features results in improved classification accuracy. Feature selection is employed to obtain the final representation and the classification problem is solved using SVM. Similarly, the Weber-normalized \cite{Chen:TPAMI10} faces are obtained in \cite{Kohli:BTAS12} to alleviate the illumination variation in the data. The features are detected around salient points, that are extracted using Difference of Gaussian. A similarity vector is then computed around each pair of salient points by employing the Self-Similarity Descriptor (SSD) \cite{Shechtman:CVPR07}. The distance measure is then classified by an SVM classifier.

On the other hand, in order to perform tri-subject kinship verification, each image is partitioned into overlapping patches and SIFT features are extracted in \cite{Qin:TransMulti15}. Feature selection is then performed and Symmetric Bilinear Model (SBM) is used to learn the similarity between the parents and the child, by learning the pairwise similarities first. The verification function is modelled as the linear combination of the two similarities. Calculating the tri-subject similarity in two steps can induce noise to the model, since the tri-person inheritance is not considered. Based on that, Zhang et al. \cite{Zhang:ICIP16} introduced a model to compute tri-subject dissimilarity. An over-complete set of features from High Dimensional LBP Histograms (HDLBPH) is employed as facial representation. The experimental results indicate the effectiveness of modeling tri-subject kinship jointly instead of using two bi-subject models.

\subsection{Subspace learning-based methods}

The methods in this subsection focus on learning a kinship-invariant subspace by making use of tools like factor analysis and transfer learning. The aforementioned descriptors can be used as local descriptors to build the subspace. In particular, LBP, LPQ and SIFT descriptors are used in \cite{Duan:ICIP15}. Based on the hypothesis that, symmetry information can capture non kinship variations like pose and illumination, the symmetry feature is employed as the kinship unrelated part of an image. The kinship unrelated part is then subtracted from the original image and a classifier performs verification. The experimental results indicate that LPQ perform better than the other descriptors on the KinFaceW-II dataset, while LBP perform better on the KinFaceW-I dataset.

Midlevel features are employed in \cite{Yan:TransCYber15}, where LBP and SIFT descriptors are used as low-level image representations. The final subspace is obtained using Prototype Discriminative Feature Learning (PDFL). For this method, two datasets are used, one with labelled kinship and one without. To obtain the subspace, the objective function minimizes the difference between kin-related samples and maximizes the difference between neighbouring non-kin samples. An SVM classifier is used for verification, while the experimental results indicate improved performance compared to the state-of-the-art. Nevertheless, the model fails to outperform the human observer for some kinds of kinship.

Based on the assumption that parents resemble their children more closely when they are younger compared to when they are older, transfer subspace tools are used in \cite{Xia:TransMulti12} and \cite{Shao:CVPRw11} to determine kinship in photos. 
In particular, the divergence between the distributions of children and old parents is minimized, using the young parents as an intermediate set. Gabor features and distance ratios from the anthropometic model \cite{Ramanathan:CVPR06} are used as descriptors and the resulting subspace maximizes the child-young parent and child-old parent similarities. Experiments on context-aware kinship verification are performed and are presented in the subsection \ref{subS:discussKin}.

\subsection{Metric learning-based methods}

Metric Learning methods \cite{Xing:MetricLearning} aim at learning a distance metric that gets similar samples closer than dissimilar ones. A number of methods have successfully used the tools of metric learning to perform kinship verification using different facial features. In particular, Ensemble Metric Learning with sample and feature selection is employed in \cite{Somanath:BTAS12} to create a mapping that reduces the distance between images of related people. The local representations include Spatial histogram of SIFT (SH-SIFT), as well as a vector containing intensity, Pyramid Histogram of Gradients (PHOG) \cite{Bosch:ACM07} and Gabor wavelet at four scales and six orientations. The final decision is obtained using SVR. The experiments reveal the implications of age difference and gender on the performance of the kinship verification model, which are discussed in the next subsection.

In order to learn a metric that not only projects kin-related faces close but also pulls unrelated ones out of their neighbourhood, Neighbourhood Repulsed Metric Learning (NRML) is introduced in \cite{Lu:TPAMI14}. Further to that, Multiview NRML is proposed to obtain metrics for multiple feature representations and to deal with multiview data. Four different descriptors are used, namely LBP, SIFT, TPLBP and LE \cite{Cao:CVPR10}. The last one outperforms the other descriptors in the experiments, due to the fact that it is directly learned from the data. The method is extended to handle perioculal images in \cite{Patel:CVIU17}. The block-based NRML (BNRML) uses the histograms from the block pairs to learn multiple distance metrics. The method uses LTP \cite{Tan:LTP} features and outperforms the original NRML method on the KinFaceW benchmarks.

Multiple distance metrics for different sets of features are learned in \cite{Yan:TIFS14} and Discriminative Multi-Metric Learning (DMML) is introduced. Three different descriptors are employed, namely LBP, SIFT and Spatial Pyramid LEarning (SPLE) \cite{Zhou:ACM11} and the method learns a metric for each of them simultaneously. This method \cite{Yan:TIFS14} performs better for multiple feature metric learning, as well as single metric learning, while the SPLE appears to be the best descriptor for the task. In both methods, kinship is decided using an SVM classifier.

Multiple feature representations are also employed in  \cite{Hu:ACCV14} and \cite{Hu:TCSVT17}. Instead of learning a distance metric with concatenated feature vectors, Large Margin Multi-Metric Learning (LM$^{3}$L) \cite{Hu:ACCV14}, \cite{Hu:TCSVT17} is introduced and multiple distance metrics are learned jointly. In that way, more discriminative and complementary information is exploited, as the correlation of the different representations is maximized. Furthermore, to better exploit the local manifold structure of images, Local Large Margin Multi-Metric Learning  (L$^{2}$M$^{3}$L) \cite{Hu:TCSVT17} is introduced to incorporate Local Metric Learning with LM$^{3}$L.

Learning a Mahalanobis distance metric is equivalent to finding a linear transformation that projects the samples to a subspace, where the Euclidean distance of the similar samples is smaller than the dissimilar ones. 
%added after review
Similarly, a linear transformation that minimizes the correlation, instead of the euclidean distance, is learned in \cite{Yan:IMAVIS17}.
%added after review
In order to learn a set of non-linear transformations, Discriminative Deep Metric Learning (DDML) is introduced in \cite{Lu:TIP17}. This method uses neural networks to project the faces to a discriminative subspace and is also extended to Descriminative Deep Multi Metric Learning (DDMML) for multiple features. The experimental results show that deep architectures that employ multiple features perform better for the tasks of face and kinship verification. Subsection \ref{subS:deep kin} features more on deep hierarchical representations for kinship verification.

\subsection{Deep learning-based representations} \label{subS:deep kin}

Similar to the aforementioned tasks in this survey, Deep Learning methodologies have shown some of the most promising results for kinship verification. The widely used architectures, e.g., VGG16, have been used for this task \cite{Robinson:FIW} to exploit large datasets like the Families In the Wild dataset. More sophisticated deep learning methods introduced for kinship verification are described as follows.

A deep autoencoder is used in \cite{Dehgan:CVPR14} to encode the relationship between the faces in two images. Instead of encoding each image separately, a relational model that uses two images as input is introduced. The gated autoencoder model outperforms most metric learning techniques, as it learns the features and the metrics jointly. The output is a statistic about relatedness and resemblance, which is used to perform kinship verification. The experimental results indicate the genetic inheritance of facial features in the Family 101 dataset. In particular, sons appear to resemble more their fathers, while the opposite stands for the daughters.

One of the latest works in kinship verification is the filtered contractive Deep Belief Network (fcDBN) proposed in \cite{Kohli:TIP16}. A DBN  is a deep learning method that consists of multiple stacked Restricted Boltzman Machines (RBM) \cite{Hinton:DBN}. The deep representations are obtained by applying filters to capture the inherent structure of the images.  In order to train the fcDBN model a large number of images from multiple datasets are used, a strategy that had not been previously employed for this task before. Kinship verification is performed by a 3-layer Neural Network and the method succeeds in outperforming the state-of-the-art methods in multiple datasets.

\subsection{Other representations}

Additionally to the above, other approaches have been introduced for the task of Kinship verification. Inheritable genetic transformation is modelled in \cite{Liu:BTAS15} and \cite{Puthen:ICIP16}. In particular, Fisher Vector \cite{Simonyan:BMVC13} in Opponent Colour Space are used in \cite{Puthen:ICIP16} as local representations. The inheritable information is captured by learning a common transformation on the representations of child and parent. On the other hand, the SIFT flow algorithm \cite{Liu:TPAMI11} is employed in \cite{Puthen:ICIP16} and the inheritable transformation is similarly learned so that the kin-related representations are as close as possible.

Instead of only applying texture descriptors, Geometric information is also incorporated in \cite{Wang:ICIP14}. Facial landmarks are used and the representation is obtained using the methods of the Grassmann manifold. In addition, appearance features are extracted by using LBP on a Gaussian image pyramid and are fused with the geometric features. A Gaussian Mixture Model (GMM) is used for similarity feature extraction from the appearance features. Verification is achieved using an SVM and the experiments indicate that even though geometric information is not effective on its own for kinship verification, it significantly boosts the performance when fused with appearance information.

Among other modalities, contextual and dynamic features have been studied for kinship verification. In particular, kinship verification in a photo using prior context knowledge is studied in \cite{Xia:TransMulti12}. The contextual information included captures gender relation, age difference, relative distance and kinship score. The experiments show that contextual information is useful for kinship verification. Lastly, dynamic features from smiles are employed studied in \cite{Dibeklioglu:ICCV13}. The dynamic features and are used along with Completed LBP from Three Orthogonal Patterns (CLBP-TOP). The experimental results indicate that smile dynamics are not sufficiently discriminative when used individually, but like geometric features, they can increase the accuracy of kinship verification in combination with spatio-temporal features. Interestingly, the performance of the model is higher for spontaneous smiles rather than posed ones.

\subsection{Evaluation protocols, metrics, and results for kinship verification}

\textit{\textbf{Protocols: }}
The most widely adopted Evaluation Protocol for Kinship Verification is k-fold cross-validation, particularly 5-fold cross-validation (5-fold c-v). This protocol is used for most of the datasets in this subsection. Specifically for the Vadana dataset, a different protocol is proposed. The dataset is split into 6 subsets, 3 including parent-child (VADANA-PC) and 3 including sibling (VADANA-S). For each relationship, two subsets included only adult faces and one both adult and child faces. Lastly, the aforementioned leave-one-out protocol protocol is used for UvA-NEMO dataset.

\textit{\textbf{Metrics: }}
The common evaluation metric for the task of kinship verification is the classification accuracy (acc.). The results are usually obtained for every subset, indicating a different kinship relation, as well as on the whole dataset. In cases where the accuracy is obtained on more than one set, instead of the average, we include the best and the worst scores reported. Aforementioned metrics, like EER, are also applied to kinship verification. The metric used in the VADANA protocol is the accuracy at EER (AEER).

\textit{\textbf{Results:}} 
The best reported results for each dataset, along with the evaluation protocols are tabulated in Table \ref{T:small_kin}. The methods in this subsection are presented in chronological order in Table \ref{T:large kin}.

\begin{table}[htbp]
  \center
  \caption{Best reported results on Kinship Verification (the results in \cite{Guo:ICPR14} are for Kinship Recognition)}
    \begin{tabular}{lllllllr}
    \toprule
    dataset& method& metric & protocol &score \\
          \midrule
    TSKinFace&\cite{Zhang:ICIP16} & acc. & 5-fold c-v& 89.8\% \\
    CornellKin&\cite{Kohli:TIP16} & acc.& 5-fold c-v& 89.5\%  \\
    FIW&\cite{Robinson:FIW} & acc.& 5-fold c-v& 71\% $\pm 2.3$  \\
    WVU KInship&\cite{Kohli:TIP16} & acc.& 5-fold c-v& 90.8\%  \\
    VADANA&\cite{Gong:CVPR15} & AEER & vadana & 60.43-80.18\%  \\
    KinFaceW-I&\cite{Kohli:TIP16} & acc.& 5-fold c-v & 96.1\%  \\
    KinFaceW-II&\cite{Kohli:TIP16} & acc.& 5-fold c-v & 96.2\% \\
    Family 101&\cite{Wang:ICIP14}& acc.&5-fold c-v & 92.03\%  \\
    UvA-NEMO&\cite{Dibeklioglu:ICCV13} & acc.&  leave-one-out& 67.11\% \\
    Sibling-Face&\cite{Guo:ICPR14} & acc.& N/A & 52.48\% \\
    Group-Face&\cite{Guo:ICPR14} & acc. & N/A & 69.25\% \\
    UBKinFace&\cite{Kohli:TIP16} & acc. & 5-fold c-v& 91.8\% \\
    IIITD-Kinship&\cite{Kohli:BTAS12} & acc. & 5-fold c-v& 75.2\% \\

    \bottomrule
    \end{tabular}%
  \label{T:small_kin}%
\end{table}%

\subsection{Discussion on kinship verification} \label{subS:discussKin}

The performance of kinship verification models can be affected by a number of factors. Some of the factors that can deteriorate the performance include pose, image quality \cite{Bottino:ICPR12}, lighting conditions \cite{Qin:TransMulti15}, as well as gender and age difference between kin \cite{Somanath:BTAS12}. Experiments are conducted in the cited studies to define the effect of each factor.

In order to determine the effectiveness of the machine in kinship verification, a study on kinship verification by humans is performed in \cite{Lu:TPAMI14}. The results indicate that humans perform slightly worse than machines, while contextual information such as hair, colour and background can improve their accuracy. Moreover, a study of the several factors affecting human accuracy at verifying kin \cite{Kohli:TIP16} indicates that women outperform men, while older people can distinguish kin better. The kin relationships containing females appear also to be detected more accurately, which may be attributed occlusions like beard or mustache. The study also explains how much kin-related information is captured in different face patches.

\subsection{Challenges in the wild}

With the oldest benchmark dating back to 2010, kinship verification is a field in its infancy. As mentioned in Section \ref{S:data}, the majority of methods in this subsection is evaluated on in the wild data. Nevertheless, Table \ref{T:small_kin} indicates significant differences between the state-of-the-art performance on each dataset (e.g., 96.2\% on KinFaceW II and 71\% on FIW). These differences can be interpreted by the bias in experimental set ups inflicted by gathering images from the same photo \cite{Lopez:TPAMI16}. Additionally, the laborious nature of gathering these images has resulted in small datasets that do not contain extreme variation with regards to pose, occlusions or image quality. On the other hand, the much larger FIW dataset demonstrates larger variation in pose, image quality as well as age of the subjects. The problem of kinship verification in the wild has been tackled in two competitions \cite{Lu:kincomp14},\cite{Lu:kincomp15}. The second is based on the KinFaceW datasets while a new dataset was collected for the first one. The difference in the results indicates that the KinFaceW datasets are indeed less challenging.

\begin{table*}[]
\centering
\begin{tabular}{llllll}
\textbf{Year} & \textbf{Paper}                     & \textbf{Representation}                                               & \textbf{Method}                                 & \textbf{Dataset(score)}           & \textbf{Metric} \\
\midrule
2010 & \cite{Fang:ICIP10}        & FP, FD, Colour, GH & kNN, SVM                               & CornellKin(70.67\%)      & acc.   \\
\midrule
2011 & \cite{Shao:CVPRw11}       & Gabor                                               & CMC,TSL, kNN                           & UBKinFace(56.5\%)        & acc.   \\
\midrule
2012 & \cite{Guo:TIM12}         & DAISY                                               & Bayes decision                         & private(75\%)            & acc.   \\
2012 & \cite{Xia:ICPR12}         & binary, relative attributes                         & SVM                                    & UBKinFace(82.5\%)        & acc.   \\
2012 & \cite{Kohli:BTAS12}           & DoG salient points                                              & SSD, SVM                               & IITD(75.2\%)             & acc.   \\
2012 & \cite{Xia:TransMulti12}   & Gabor                                               & CMC, TSL, kNN                          & UBKinFace(56.5\%)        & acc.   \\
2012 & \cite{Somanath:BTAS12}    & PHOG, Gabor, SH-SIFT                                & Meric Learning, SVR                    & VADANA-PC(80.18\%)       & AEER   \\
     &                           &                                                     &                                        & VADANA-S(75.64\%)        & AEER   \\
\midrule
2013 & \cite{Fang:ICIP13}        & dSIFT                                               & Sparse Group Lasso                     & Family101(32\%)          & acc.   \\
2013 & \cite{Dibeklioglu:ICCV13} & dynamic, CLBP-TOP                                   & mRMR, SVM                              & UvA-NEMO(67.11\%)        & acc.   \\
\midrule
2014 & \cite{Lu:TPAMI14}         & LBP, SIFT, TPLBP and LE                             & MNRLM, SVM                             & KinFaceW-I(69.9\%)       & acc.   \\
     &                           &                                                     &                                        & KinFaceW-II(76.5\%)      & acc.   \\
2014 & \cite{Yan:TIFS14}         & LBP, SIFT, SPLE                                     & DMML, SVM                              & KinFaceW-I(69.5-75.5\%)  & acc.   \\
     &                           &                                                     &                                        & KinFaceW-II(76.5-79.5\%) & acc.   \\
     &                           &                                                     &                                        & CornellKin(70.5-77.5\%)  & acc.   \\
     &                           &                                                     &                                        & UBKinFace(70-74.5\%)     & acc.   \\
2014 & \cite{Hu:ACCV14}          & LE, LBP, TPLBP, SIFT                                & LMMML                                  & KinFaceW-II(78.7\%)      & acc.   \\
2014 & \cite{Dehgan:CVPR14}      & PI                                                  & gated autoencoder                      & KinFaceW-I(74.5\%)       & acc.   \\
     &                           &                                                     &                                        & KinFaceW-II(82.5\%)      & acc.   \\
2014 & \cite{Wang:ICIP14}        & LBP, Grassman Manifold                              & GMM, geodesic distance, SVM            & Family101(92.03\%)       & acc.   \\
2014 & \cite{Guo:ICPR14}         & LBP,BIF, relative age                               & logistic regression,  graph classifier & Custom(52.48\%)         & acc.   \\
     &                           &                                                     &                                        & GroupFace(69.25\%)       & acc.   \\
\midrule
2015 & \cite{Bottinok:FG15}      & LPQ, TPLBP, FPLBP, WLD                              & mRMR, SFFS, SVM                        & KinFaceW-I(86.3\%)       & acc.   \\
     &                           &                                                     &                                        & KinFaceW-II(83.1\%)      & acc.   \\
2015 & \cite{Qin:TransMulti15}   & SIFT                                                & RSBM                                   & TSKinFace(85.4\%)        & acc.   \\
2015 & \cite{Duan:ICIP15}        & LPQ                                                 & PCA, feature substraction              & KinFaceW-I(70.9\%)       & acc.   \\
     &                           &                                                     &                                        & KinFaceW-II(77.1\%)      & acc.   \\
2015 & \cite{Yan:TransCYber15}   & LBP, SIFT                                           & PDFL, SVM                              & KinFaceW-I(70.1\%)       & acc.   \\
     &                           &                                                     &                                        & KinFaceW-II(77\%)        & acc.   \\
     &                           &                                                     &                                        & CornellKin(71.9\%)       & acc.   \\
     &                           &                                                     &                                        & UBKinFace(67.3\%)        & acc.   \\
\midrule
2016 & \cite{Zhang:ICIP16}       & HDLBP                                               & dissimilarity vector, mRMR             & TSKinFace(89.7\%)        & acc.   \\
\midrule
2017 & \cite{Hu:TCSVT17}         & LE, LBP, TPLBP, SIFT                                & LLMMML                                 & KinFaceW-II(80\%)        & acc.   \\
2017 & \cite{Lu:TIP17}           & LBP,DSIFT, HOG, LPQ                                 & DDMML                                  & KinFaceW-I(83.5\%)       & acc.   \\
     &                           &                                                     &                                        & KinFaceW-II(84.3\%)      & acc.   \\
     &                           &                                                     &                                        & TSKinFace(82.5-88.5\%)   & acc.   \\
2017 & \cite{Yan:IMAVIS17}                & LE                                                & NRCML                                    & KinFaceW-I(66.3\%)        & acc.   \\
     &                           &                                                     &                                        & KinFaceW-II(78.8\%)        & acc.   \\
2017 & \cite{Patel:CVIU17}                & LTP                                              & BNRML                                   & KinFaceW-I(78.7\%)        & acc.   \\
     &                           &                                                     &                                        & KinFaceW-II(80.55\%)        & acc.   \\
2017 & \cite{FIW}                & PI                                                  & VGG                                    & FIW(71\%$\pm{2.3}$)        & acc.   \\
2017 & \cite{Kohli:TIP16}        & PI                                                  & fcDBN,3-l NN                           & KinFaceW-I(96.1\%)       & acc.   \\
     &                           &                                                     &                                        & KinFaceW-II(96.2\%)      & acc.   \\
     &                           &                                                     &                                        & CornellKin(89.5\%)       & acc.   \\
     &                           &                                                     &                                        & UBKinFace(91.8\%)        & acc.   \\
     &                           &                                                     &                                        & WVU(90.8\%)              & acc.  
\end{tabular}
\caption{Overview of Kinship Verification Methods}
\label{T:large kin}
\end{table*}

\section{Aging and Kinship}

The fact that both aging and kinship are genetically encoded \cite{Browner:2004} indicates an inherent synergy between the two. This synergy is used in \cite{Shu:PR16} to perform kinship guided age progression. The parent face serves as a prior to predict the aging of the child. The kinship information is incorporated by morphing the age-progressed face to the one of the parent. On the other hand, old and young parent face images are used in \cite{xia:ub1}. The correlation between the aging of the parent face and their children is leveraged by a transfer learning model from the young parent-child to the old parent-child domain.  

Besides the possible synergy between them, aging and kinship modeling face similar issues when it comes to dealing with faces in the wild. Variations like illumination, pose, expression and image quality deteriorate the accuracy of such systems. The methods in the previous sections apply classical face analysis descriptors, e.g., LBP, SIFT, HOG, as well as state-of-the-art deep CNN to attack these problems. Furthermore, aging and kinship modeling share a number of protocols. In particular, cross-age face verification and kinship verification systems aim to learn a similarity metric and similar methods can be applied. This approach is adopted in \cite{Lu:TPAMI14}, where the task of cross-age face verification is tackled as 'self-kinship verification' using metric learning.

\section{Conclusions}\label{subS:Conc}

Motivated by the increasing interest and plethora of real world applications, the state-of-the-art methods in age and kinship modeling have been surveyed. The main challenges and results of the methods are described for each task individually. The results tabulated in Tables 4, 7, 9 and 11 indicate the superiority of deep learning methods in modeling the non-linear transformations of aging and kinship. Specifically, the discriminative power of deep learning representations, in addition to the availability of in the wild image datasets, have produced results that surpass human abilities in the tasks of age estimation, age-invariant facial characterization and kinship verification. On the other hand, the application of deep generative models (e.g., GANs) for the task of age progression has produced photorealistic age-progressed images with impressive high-frequency details.

Nevertheless, deep learning (DL) methods demonstrate a number of limitations that hinder the effective use of such methods for modeling complex transformations like aging and kinship. To approximate such functions DL methods use hierarchical models that need sufficiently dense sampling from the data distribution, that is, they are data hungry. The resulting models are highly non-linear and complex, creating systems that are not mathematically transparent, while being difficult to train with no guarantee of convergence. This deems DL methods unsuitable for critical tasks due to their lack of interpretability and sensitivity to adversarial examples \cite{Goodfellow:adv15}.

To further advance the fields of aging and kinship modeling, the community should focus on diversifying the modalities of the data, since the vast majority of research focuses on 2-Dimensional data. Although some papers indicate the possibility of using dynamic representation for these tasks (e.g., \cite{Dibeklioglu:ICCV13}, \cite{Dibekliouglu:TIP15}), the lack of labelled videos captured in the wild delays progress in this problem. Lastly, the lack of large in the wild datasets with multiple annotations, i.e. both age and kinship, does not allow for research on the correlation between age and kinship. Since both kinship and aging transformations are genetically encoded, the inheritance of aging patterns can be investigated. Such datasets will also  reveal the possibility of incorporating age or kinship as soft-biometrics to boost the accuracy of face recognition systems (e.g., in \cite{Kohli:TIP16}) as well as investigate the synergies between the two.

\bibliographystyle{abbrv}
{\footnotesize{
\bibliography{survey}

\begin{thebibliography}{100}

\bibitem{LHI}
L{HI} {I}mage {D}atabase.
\newblock 2010.

\bibitem{Abate:2d3dsurvey}
A.~F. Abate, M.~Nappi, D.~Riccio, and G.~Sabatino.
\newblock 2d and 3d face recognition: A survey.
\newblock {\em Pattern recognition letters}, 28(14):1885--1906, 2007.

\bibitem{Agustsson:APPAREAL}
E.~Agustsson, R.~Timofte, S.~Escalera, X.~Baro, I.~Guyon, and R.~Rothe.
\newblock Apparent and real age estimation in still images with deep residual
  regressors on appa-real database.
\newblock In {\em IEEE International Conference on Automatic Face \& Gesture
  Recognition (FG)}, 2017.

\bibitem{Ahonen:TPAMI06}
T.~Ahonen, A.~Hadid, and M.~Pietikainen.
\newblock Face description with local binary patterns: Application to face
  recognition.
\newblock {\em IEEE transactions on pattern analysis and machine intelligence},
  28(12):2037--2041, 2006.

\bibitem{Alley:1988}
R.~A. Alley.
\newblock {\em Social and Applied Aspects of Perceiving Faces}.
\newblock Erlbaum, Hillsdale, NJ, 1988.

\bibitem{Alnajar:ICIP15}
F.~Alnajar, T.~Gevers, and S.~Karaoglu.
\newblock Age estimation under changes in image quality: An experimental study.
\newblock In {\em IEEE International Conference on Image Processing (ICIP)},
  pages 3987--3991, Sept 2015.

\bibitem{Antipov:CVPRw16}
G.~{A}ntipov, M.~{B}accouche, S.-A. {B}errani, and J.-L. {D}ugelay.
\newblock {A}pparent age estimation from face images combining general and
  children-specialized deep learning models.
\newblock In {\em {IEEE} {C}onference on {C}omputer {V}ision and {P}attern
  {R}ecognition {W}orkshops (CVPR-W)}, 2016.

\bibitem{Antipov:ICIP17}
G.~Antipov, M.~Baccouche, and J.-L. Dugelay.
\newblock Face aging with conditional generative adversarial networks.
\newblock {\em IEEE International Conference on Image Proccessing (ICIP)},
  2017.

\bibitem{Bar:ICML03}
A.~Bar-Hillel, T.~Hertz, N.~Shental, and D.~Weinshall.
\newblock Learning distance functions using equivalence relations.
\newblock In {\em Proceedings of the 20th International Conference on Machine
  Learning (ICML)}, pages 11--18, 2003.

\bibitem{IranianFace}
A.~Bastanfard, M.~A. Nik, and M.~M. Dehshibi.
\newblock Iranian face database with age, pose and expression.
\newblock In {\em Machine Vision, 2007. ICMV 2007. International Conference
  on}, pages 50--55. IEEE, 2007.

\bibitem{Bauckhage:ICPR10}
C.~Bauckhage, A.~Jahanbekam, and C.~Thurau.
\newblock Age recognition in the wild.
\newblock In {\em International Conference on Pattern Recognition (ICPR)},
  pages 392--395, 2010.

\bibitem{Bay:SURF}
H.~Bay, T.~Tuytelaars, and L.~Van~Gool.
\newblock Surf: Speeded up robust features.
\newblock {\em European Conference onComputer vision (ECCV)}, pages 404--417,
  2006.

\bibitem{Bereta:PR13}
M.~Bereta, P.~Karczmarek, W.~Pedrycz, and M.~Reformat.
\newblock Local descriptors in application to the aging problem in face
  recognition.
\newblock {\em Pattern Recognition}, 46(10):2634--2646, 2013.

\bibitem{Bhattarai:ICASSP16}
B.~Bhattarai, G.~Sharma, A.~Lechervy, and F.~Jurie.
\newblock A joint learning approach for cross domain age estimation.
\newblock In {\em IEEE International Conference on Acoustics, Speech and Signal
  Processing (ICASSP)}, pages 1901--1905, March 2016.

\bibitem{Bienenstock:NeuronSelect}
E.~L. Bienenstock, L.~N. Cooper, and P.~W. Munro.
\newblock Theory for the development of neuron selectivity: orientation
  specificity and binocular interaction in visual cortex.
\newblock Technical report, DTIC Document, 1981.

\bibitem{Biswas:BTAS08}
S.~Biswas, G.~Aggarwal, N.~Ramanathan, and R.~Chellappa.
\newblock A non-generative approach for face recognition across aging.
\newblock In {\em IEEE International Conference on Biometrics: Theory,
  Applications and Systems (BTAS)}, pages 1--6. IEEE, 2008.

\bibitem{Bosch:ACM07}
A.~Bosch, A.~Zisserman, and X.~Munoz.
\newblock Representing shape with a spatial pyramid kernel.
\newblock In {\em Proceedings of the 6th ACM international conference on Image
  and video retrieval}, pages 401--408. ACM, 2007.

\bibitem{Bottino:ICPR12}
A.~G. Bottino, M.~De~Simone, A.~Laurentini, and T.~Vieira.
\newblock A new problem in face image analysis: finding kinship clues for
  siblings pairs.
\newblock 2012.

\bibitem{Bottinok:FG15}
A.~Bottinok, I.~U. Islam, and T.~F. Vieira.
\newblock A multi-perspective holistic approach to kinship verification in the
  wild.
\newblock In {\em IEEE International Conference and Workshops on Automatic Face
  and Gesture Recognition (FG)}, volume~2, pages 1--6. IEEE, 2015.

\bibitem{Bouchaffra:TNN15}
D.~Bouchaffra.
\newblock Nonlinear topological component analysis: Application to
  age-invariant face recognition.
\newblock {\em IEEE Transactions on Neural Networks and Learning Systems},
  26(7):1375--1387, July 2015.

\bibitem{Breiman:Bagging}
L.~Breiman.
\newblock Bagging predictors.
\newblock {\em Machine learning}, 24(2):123--140, 1996.

\bibitem{Browner:2004}
W.~S. Browner, A.~J. Kahn, E.~Ziv, A.~P. Reiner, J.~Oshima, R.~M. Cawthon,
  W.-C. Hsueh, and S.~R. Cummings.
\newblock The genetics of human longevity.
\newblock {\em The American journal of medicine}, 117(11):851--860, 2004.

\bibitem{Bruce:98}
V.~Bruce and A.~Young.
\newblock {\em In the eye of the beholder: the science of face perception.}
\newblock Oxford University Press, 1998.

\bibitem{Burt:Caucasian}
D.~M. Burt and D.~I. Perrett.
\newblock Perception of age in adult caucasian male faces: Computer graphic
  manipulation of shape and colour information.
\newblock {\em Proceedings of the Royal Society of London B: Biological
  Sciences}, 259(1355):137--143, 1995.

\bibitem{Cai:TIP06}
D.~Cai, X.~He, J.~Han, and H.-J. Zhang.
\newblock Orthogonal laplacianfaces for face recognition.
\newblock {\em IEEE transactions on image processing}, 15(11):3608--3614, 2006.

\bibitem{Cao:CVPR10}
Z.~Cao, Q.~Yin, X.~Tang, and J.~Sun.
\newblock Face recognition with learning-based descriptor.
\newblock In {\em IEEE Conference on Computer Vision and Pattern Recognition
  (CVPR)}, pages 2707--2714. IEEE, 2010.

\bibitem{Chang:TIP15}
K.~Y. Chang and C.~S. Chen.
\newblock A learning framework for age rank estimation based on face images
  with scattering transform.
\newblock {\em IEEE Transactions on Image Processing}, 24(3):785--798, March
  2015.

\bibitem{Chang:ICPR10}
K.~Y. Chang, C.~S. Chen, and Y.~P. Hung.
\newblock A ranking approach for human ages estimation based on face images.
\newblock In {\em International Conference on Pattern Recognition (ICPR)},
  pages 3396--3399, 2010.

\bibitem{Chang:CVPR11}
K.~Y. Chang, C.~S. Chen, and Y.~P. Hung.
\newblock Ordinal hyperplanes ranker with cost sensitivities for age
  estimation.
\newblock In {\em IEEE Conference on Computer Vision and Pattern Recognition
  (CVPR)}, pages 585--592, 2011.

\bibitem{Chao:PR13}
W.-L. Chao, J.-Z. Liu, and J.-J. Ding.
\newblock Facial age estimation based on label-sensitive learning and
  age-oriented regression.
\newblock {\em Pattern Recognition}, 46(3):628--641, 2013.

\bibitem{Chen:ECCV14}
B.-C. Chen, C.-S. Chen, and W.~H. Hsu.
\newblock Cross-age reference coding for age-invariant face recognition and
  retrieval.
\newblock In {\em European Conference on Computer Vision (ECCV)}, 2014.

\bibitem{Chen:TransMult15}
B.~C. Chen, C.~S. Chen, and W.~H. Hsu.
\newblock Face recognition and retrieval using cross-age reference coding with
  cross-age celebrity dataset.
\newblock {\em IEEE Transactions on Multimedia}, 17(6):804--815, June 2015.

\bibitem{Chen:FG11}
C.~Chen, W.~Yang, Y.~Wang, K.~Ricanek, and K.~Luu.
\newblock Facial feature fusion and model selection for age estimation.
\newblock In {\em IEEE International Conference and Workshops on Automatic Face
  and Gesture Recognition (FG)}, pages 200--205, March 2011.

\bibitem{Chen:CVPR00}
H.~F. Chen, P.~N. Belhumeur, and D.~W. Jacobs.
\newblock In search of illumination invariants.
\newblock In {\em Computer Vision and Pattern Recognition, 2000. Proceedings.
  IEEE Conference on}, volume~1, pages 254--261. IEEE, 2000.

\bibitem{Chen:TPAMI10}
J.~Chen, S.~Shan, C.~He, G.~Zhao, M.~Pietikainen, X.~Chen, and W.~Gao.
\newblock Wld: A robust local image descriptor.
\newblock {\em IEEE transactions on pattern analysis and machine intelligence},
  32(9):1705--1720, 2010.

\bibitem{Chen:WACV16}
J.-C. Chen, V.~M. Patel, and R.~Chellappa.
\newblock Unconstrained face verification using deep cnn features.
\newblock In {\em IEEE Winter Conference on Applications of Computer Vision
  (WACV)}, pages 1--9. IEEE, 2016.

\bibitem{Chen:CVPR13}
K.~Chen, S.~Gong, T.~Xiang, and C.~C. Loy.
\newblock Cumulative attribute space for age and crowd density estimation.
\newblock In {\em IEEE Conference on Computer Vision and Pattern Recognition},
  pages 2467--2474, June 2013.

\bibitem{Chen:CVPR17}
S.~Chen, C.~Zhang, M.~Dong, J.~Le, and M.~Rao.
\newblock Using ranking-cnn for age estimation.

\bibitem{Chen:TIFS13}
Y.~L. Chen and C.~T. Hsu.
\newblock Subspace learning for facial age estimation via pairwise age ranking.
\newblock {\em IEEE Transactions on Information Forensics and Security},
  8(12):2164--2176, Dec 2013.

\bibitem{Cootes:AAM}
T.~F. Cootes, G.~J. Edwards, and C.~J. Taylor.
\newblock Active appearance models.
\newblock {\em IEEE Transactions on pattern analysis and machine intelligence},
  23(6):681--685, 2001.

\bibitem{Cootes:CVIP95}
T.~F. Cootes, C.~J. Taylor, D.~H. Cooper, and J.~Graham.
\newblock Active shape models-their training and application.
\newblock {\em Computer vision and image understanding}, 61(1):38--59, 1995.

\bibitem{Cristinacce:CLM}
D.~Cristinacce and T.~F. Cootes.
\newblock Feature detection and tracking with constrained local models.
\newblock In {\em BMVC}, volume~1, page~3, 2006.

\bibitem{Burt:Biological95}
D.~I.~P. D.~Michael~Burt.
\newblock Perception of age in adult caucasian male faces: Computer graphic
  manipulation of shape and colour information.
\newblock {\em Proceedings: Biological Sciences}, 259(1355):137--143, 1995.

\bibitem{Dalal:HOG}
N.~Dalal and B.~Triggs.
\newblock Histograms of oriented gradients for human detection.
\newblock In {\em Computer Vision and Pattern Recognition, 2005. CVPR 2005.
  IEEE Computer Society Conference on}, volume~1, pages 886--893. IEEE, 2005.

\bibitem{Dantcheva:TIFS16}
A.~Dantcheva, P.~Elia, and A.~Ross.
\newblock What else does your biometric data reveal? a survey on soft
  biometrics.
\newblock {\em IEEE Transactions on Information Forensics and Security},
  11(3):441--467, 2016.

\bibitem{Dehgan:CVPR14}
A.~Dehghan, E.~G. Ortiz, R.~Villegas, and M.~Shah.
\newblock Who do i look like? determining parent-offspring resemblance via
  gated autoencoders.
\newblock In {\em Proceedings of the IEEE Conference on Computer Vision and
  Pattern Recognition (CVPR)}, pages 1757--1764, 2014.

\bibitem{Dibeklioglu:ICCV13}
H.~Dibeklioglu, A.~Ali~Salah, and T.~Gevers.
\newblock Like father, like son: Facial expression dynamics for kinship
  verification.
\newblock In {\em IEEE International Conference on Computer Vision (ICCV)},
  pages 1497--1504, 2013.

\bibitem{Dibekliouglu:TIP15}
H.~Dibeklio{\u{g}}lu, F.~Alnajar, A.~A. Salah, and T.~Gevers.
\newblock Combining facial dynamics with appearance for age estimation.
\newblock {\em IEEE Transactions on Image Processing}, 24(6):1928--1943, 2015.

\bibitem{Dibekliouglu:ECCV12}
H.~Dibeklio{\u{g}}lu, A.~Salah, and T.~Gevers.
\newblock Are you really smiling at me? spontaneous versus posed enjoyment
  smiles.
\newblock {\em European Conference on Computer Vision (ECCV)}, pages 525--538,
  2012.

\bibitem{Ding:ACM16}
C.~Ding and D.~Tao.
\newblock A comprehensive survey on pose-invariant face recognition.
\newblock {\em ACM Transactions on intelligent systems and technology (TIST)},
  7(3):37, 2016.

\bibitem{Dingman:1964}
R.~O. Dingman and P.~Natvig.
\newblock {\em Surgery of facial fractures}.
\newblock Saunders, 1964.

\bibitem{Du:CVPR15}
L.~Du and H.~Ling.
\newblock Cross-age face verification by coordinating with cross-face age
  verification.
\newblock In {\em IEEE Conference on Computer Vision and Pattern Recognition
  (CVPR)}, pages 2329--2338, June 2015.

\bibitem{Duan:ICIP15}
X.~Duan and Z.-H. Tan.
\newblock A feature subtraction method for image based kinship verification
  under uncontrolled environments.
\newblock In {\em IEEE International Conference on Image Processing (ICIP)},
  pages 1573--1577. IEEE, 2015.

\bibitem{Duong:CVPR15}
C.~N. Duong, K.~Luu, K.~G. Quach, and T.~D. Bui.
\newblock Beyond principal components: Deep boltzmann machines for face
  modeling.
\newblock In {\em 2015 IEEE Conference on Computer Vision and Pattern
  Recognition (CVPR)}, pages 4786--4794, June 2015.

\bibitem{Ebner:2010}
N.~C. Ebner, M.~Riediger, and U.~Lindenberger.
\newblock {FACES}--{A} database of facial expressions in young, middle-aged,
  and older women and men: Development and validation.
\newblock {\em Behavior research methods}, 42(1):351--362, 2010.

\bibitem{Edelman:98}
A.~Edelman, T.~A. Arias, and S.~T. Smith.
\newblock The geometry of algorithms with orthogonality constraints.
\newblock {\em SIAM journal on Matrix Analysis and Applications},
  20(2):303--353, 1998.

\bibitem{Eidinger:TIFS14}
E.~Eidinger, R.~Enbar, and T.~Hassner.
\newblock Age and gender estimation of unfiltered faces.
\newblock {\em IEEE Transactions on Information Forensics and Security},
  9(12):2170--2179, Dec 2014.

\bibitem{Ekman:77}
P.~Ekman and W.~V. Friesen.
\newblock Facial action coding system.
\newblock 1977.

\bibitem{Escalera:ICCVw15}
S.~Escalera, J.~Fabian, P.~Pardo, X.~Bar�, J.~Gonz�lez, H.~J. Escalante,
  D.~Misevic, U.~Steiner, and I.~Guyon.
\newblock Chalearn looking at people 2015: Apparent age and cultural event
  recognition datasets and results.
\newblock In {\em IEEE International Conference on Computer Vision Workshop
  (ICCV- W)}, pages 243--251, Dec 2015.

\bibitem{Escalera:CVPRw16}
S.~Escalera, M.~Torres~Torres, B.~Martinez, X.~Baro, H.~Jair~Escalante,
  I.~Guyon, G.~Tzimiropoulos, C.~Corneou, M.~Oliu, M.~Ali~Bagheri, and
  M.~Valstar.
\newblock Chalearn looking at people and faces of the world: Face analysis
  workshop and challenge 2016.
\newblock In {\em The IEEE Conference on Computer Vision and Pattern
  Recognition (CVPR) Workshops}, June 2016.

\bibitem{Exline:1963}
R.~V. Exline.
\newblock Explorations in the process of person perception: Visual interaction
  in relation to competition, sex, and need for affiliation.
\newblock {\em Journal of personality}, 31(1):1--20, 1963.

\bibitem{Fan:ICCV11}
N.~Fan.
\newblock Learning nonlinear distance functions using neural network for
  regression with application to robust human age estimation.
\newblock In {\em IEEE International Conference on Computer Vision (ICCV)},
  pages 249--254, Nov 2011.

\bibitem{Fang:ICPR10}
H.~Fang, P.~Grant, and M.~Chen.
\newblock Discriminant feature manifold for facial aging estimation.
\newblock In {\em International Conference on Pattern Recognition (ICPR)},
  pages 593--596, Aug 2010.

\bibitem{Fang:ICIP13}
R.~Fang, A.~C. Gallagher, T.~Chen, and A.~Loui.
\newblock Kinship classification by modeling facial feature heredity.
\newblock In {\em IEEE International Conference on Image Processing (ICIP)},
  pages 2983--2987. IEEE, 2013.

\bibitem{Fang:ICIP10}
R.~Fang, K.~D. Tang, N.~Snavely, and T.~Chen.
\newblock Towards computational models of kinship verification.
\newblock In {\em IEEE International Conference on Image Processing (ICIP)},
  pages 1577--1580. IEEE, 2010.

\bibitem{Farkas:Anthropometry}
L.~G. Farkas.
\newblock {\em Anthropometry of the Head and Face}.
\newblock Raven Pr, 1994.

\bibitem{Ersi:ICIP14}
E.~Fazl-Ersi, M.~E. Mousa-Pasandi, R.~Lagani�re, and M.~Awad.
\newblock Age and gender recognition using informative features of various
  types.
\newblock In {\em IEEE International Conference on Image Processing (ICIP)},
  pages 5891--5895, Oct 2014.

\bibitem{Felzenszwalb:IJCV05}
P.~F. Felzenszwalb and D.~P. Huttenlocher.
\newblock Pictorial structures for object recognition.
\newblock {\em International journal of computer vision}, 61(1):55--79, 2005.

\bibitem{HOIP}
S.~J. Foundation.
\newblock Human and object interaction processing (hoip) face database.
\newblock 2014.

\bibitem{Freiwald:Nature09}
W.~A. Freiwald, D.~Y. Tsao, and M.~S. Livingstone.
\newblock A face feature space in the macaque temporal lobe.
\newblock {\em Nature neuroscience}, 12(9):1187--1196, 2009.

\bibitem{Freund:Boosting}
Y.~Freund and R.~E. Schapire.
\newblock A desicion-theoretic generalization of on-line learning and an
  application to boosting.
\newblock In {\em European conference on computational learning theory}, pages
  23--37. Springer, 1995.

\bibitem{Fu:TPAMI14}
S.~Fu, H.~He, and Z.-G. Hou.
\newblock Learning race from face: A survey.
\newblock {\em IEEE transactions on pattern analysis and machine intelligence},
  36(12):2483--2509, 2014.

\bibitem{Fu:TPAMI10}
Y.~Fu, G.~Guo, and T.~S. Huang.
\newblock Age synthesis and estimation via faces: A survey.
\newblock {\em IEEE Transactions on Pattern Analysis and Machine Intelligence},
  32(11):1955--1976, Nov 2010.

\bibitem{Fu:TransMulti08}
Y.~Fu and T.~S. Huang.
\newblock Human age estimation with regression on discriminative aging
  manifold.
\newblock {\em IEEE Transactions on Multimedia}, 10(4):578--584, June 2008.

\bibitem{Fu:CVPR07}
Y.~Fu, M.~Liu, and T.~S. Huang.
\newblock Conformal embedding analysis with local graph modeling on the unit
  hypersphere.
\newblock In {\em IEEE Conference on Computer Vision and Pattern Recognition
  (CVPR)}, pages 1--6. IEEE, 2007.

\bibitem{fu:air}
Y.~Fu and N.~Zheng.
\newblock M-face: An appearance-based photorealistic model for multiple facial
  attributes rendering.
\newblock {\em IEEE Transactions on Circuits and Systems for Video technology},
  16(7):830--842, 2006.

\bibitem{Gallagher:Groups}
A.~C. Gallagher and T.~Chen.
\newblock Understanding images of groups of people.
\newblock In {\em IEEE Conference on Computer Vision and Pattern Recognition},
  pages 256--263. IEEE, 2009.

\bibitem{Gao:ICB09}
F.~Gao and H.~Ai.
\newblock Face age classification on consumer images with gabor feature and
  fuzzy lda method.
\newblock In {\em International Conference on Biometrics (ICB)}, pages
  132--141. Springer, 2009.

\bibitem{Geladi:PLS}
P.~Geladi and B.~R. Kowalski.
\newblock Partial least-squares regression: a tutorial.
\newblock {\em Analytica chimica acta}, 185:1--17, 1986.

\bibitem{Geng:ICPR14}
X.~Geng, Q.~Wang, and Y.~Xia.
\newblock Facial age estimation by adaptive label distribution learning.
\newblock In {\em International Conference on Pattern Recognition (ICPR)},
  pages 4465--4470, Aug 2014.

\bibitem{Geng:TPAMI13}
X.~Geng, C.~Yin, and Z.~H. Zhou.
\newblock Facial age estimation by learning from label distributions.
\newblock {\em IEEE Transactions on Pattern Analysis and Machine Intelligence},
  35(10):2401--2412, Oct 2013.

\bibitem{Geng:TPAMI07}
X.~Geng, Z.~H. Zhou, and K.~Smith-Miles.
\newblock Automatic age estimation based on facial aging patterns.
\newblock {\em IEEE Transactions on Pattern Analysis and Machine Intelligence},
  29(12):2234--2240, Dec 2007.

\bibitem{Gong:ICCV13}
D.~Gong, Z.~Li, D.~Lin, J.~Liu, and X.~Tang.
\newblock Hidden factor analysis for age invariant face recognition.
\newblock In {\em IEEE International Conference on Computer Vision (ICCV)},
  pages 2872--2879, Dec 2013.

\bibitem{Gong:CVPR15}
D.~Gong, Z.~Li, D.~Tao, J.~Liu, and X.~Li.
\newblock A maximum entropy feature descriptor for age invariant face
  recognition.
\newblock In {\em 2015 IEEE Conference on Computer Vision and Pattern
  Recognition (CVPR)}, pages 5289--5297, June 2015.

\bibitem{Goodfellow:NIPS14}
I.~Goodfellow, J.~Pouget-Abadie, M.~Mirza, B.~Xu, D.~Warde-Farley, S.~Ozair,
  A.~Courville, and Y.~Bengio.
\newblock Generative adversarial nets.
\newblock In {\em Advances in neural information processing systems (NIPS)},
  pages 2672--2680, 2014.

\bibitem{Goodfellow:adv15}
I.~Goodfellow, J.~Shlens, and C.~Szegedy.
\newblock Explaining and harnessing adversarial examples.
\newblock 2015.

\bibitem{Graves:LSTM}
A.~Graves.
\newblock Supervised sequence labelling.
\newblock In {\em Supervised Sequence Labelling with Recurrent Neural
  Networks}, pages 5--13. Springer, 2012.

\bibitem{Guo:TIP08}
G.~Guo, Y.~Fu, C.~R. Dyer, and T.~S. Huang.
\newblock Image-based human age estimation by manifold learning and locally
  adjusted robust regression.
\newblock {\em IEEE Transactions on Image Processing}, 17(7):1178--1188, 2008.

\bibitem{Guo:CVPRw10}
G.~Guo and G.~Mu.
\newblock Human age estimation: What is the influence across race and gender?
\newblock In {\em IEEE Conference on Computer Vision and Pattern Recognition
  Workshop (CVPR- W)}, pages 71--78, June 2010.

\bibitem{Guo:CVPR11}
G.~Guo and G.~Mu.
\newblock Simultaneous dimensionality reduction and human age estimation via
  kernel partial least squares regression.
\newblock In {\em IEEE Conference on Computer Vision and Pattern Recognition
  (CVPR)}, pages 657--664, June 2011.

\bibitem{Guo:CVPR09}
G.~Guo, G.~Mu, Y.~Fu, and T.~S. Huang.
\newblock Human age estimation using bio-inspired features.
\newblock In {\em IEEE Conference on Computer Vision and Pattern Recognition
  (CVPR)}, pages 112--119, June 2009.

\bibitem{Guo:ICPR10}
G.~Guo, G.~Mu, and K.~Ricanek.
\newblock Cross-age face recognition on a very large database: The performance
  versus age intervals and improvement using soft biometric traits.
\newblock In {\em International Conference on Pattern Recognition (ICPR)},
  pages 3392--3395, Aug 2010.

\bibitem{Guo:TIM12}
G.~Guo and X.~Wang.
\newblock Kinship measurement on salient facial features.
\newblock {\em IEEE Transactions on Instrumentation and Measurement},
  61(8):2322--2325, 2012.

\bibitem{Guo:CVPR12}
G.~Guo and X.~Wang.
\newblock A study on human age estimation under facial expression changes.
\newblock In {\em IEEE Conference on Computer Vision and Pattern Recognition
  (CVPR)}, pages 2547--2553, June 2012.

\bibitem{Guo:CVPR14}
G.~Guo and C.~Zhang.
\newblock A study on cross-population age estimation.
\newblock In {\em IEEE Conference on Computer Vision and Pattern Recognition
  (CVPR)}, pages 4257--4263, June 2014.

\bibitem{Guo:ICPR14}
Y.~Guo, H.~Dibeklioglu, and L.~van~der Maaten.
\newblock Graph-based kinship recognition.
\newblock In {\em International Conference on Pattern Recognition (ICPR)},
  pages 4287--4292. Citeseer, 2014.

\bibitem{Han:TPAMI15}
H.~Han, C.~Otto, X.~Liu, and A.~K. Jain.
\newblock Demographic estimation from face images: Human vs. machine
  performance.
\newblock {\em IEEE Transactions on Pattern Analysis and Machine Intelligence},
  37(6):1148--1161, June 2015.

\bibitem{He:ResNet}
K.~He, X.~Zhang, S.~Ren, and J.~Sun.
\newblock Deep residual learning for image recognition.
\newblock In {\em IEEE Conference on Computer Vision and Pattern Recognition
  (CVPR)}, pages 770--778, 2016.

\bibitem{He:TPAMI05}
X.~He, S.~Yan, Y.~Hu, P.~Niyogi, and H.-J. Zhang.
\newblock Face recognition using laplacianfaces.
\newblock {\em IEEE transactions on pattern analysis and machine intelligence},
  27(3):328--340, 2005.

\bibitem{Hinton:DBN}
G.~E. Hinton, S.~Osindero, and Y.-W. Teh.
\newblock A fast learning algorithm for deep belief nets.
\newblock {\em Neural computation}, 18(7):1527--1554, 2006.

\bibitem{Hu:TCSVT17}
J.~Hu, J.~Lu, Y.-P. Tan, J.~Yuan, and J.~Zhou.
\newblock Local large-margin multi-metric learning for face and kinship
  verification.
\newblock {\em IEEE Transactions on Circuits and Systems for Video Technology},
  2017.

\bibitem{Hu:ACCV14}
J.~Hu, J.~Lu, J.~Yuan, and Y.-P. Tan.
\newblock Large margin multi-metric learning for face and kinship verification
  in the wild.
\newblock In {\em Asian Conference on Computer Vision}, pages 252--267.
  Springer, 2014.

\bibitem{Hu:TIP17}
Z.~Hu, Y.~Wen, J.~Wang, M.~Wang, R.~Hong, and S.~Yan.
\newblock Facial age estimation with age difference.
\newblock {\em IEEE Transactions on Image Processing}, 26(7):3087--3097, 2017.

\bibitem{Huang:DenseMet}
G.~Huang, Z.~Liu, K.~Q. Weinberger, and L.~van~der Maaten.
\newblock Densely connected convolutional networks.
\newblock In {\em IEEE Conference on Computer Vision and Pattern Recognition
  (CVPR)}, 2017.

\bibitem{LFW}
G.~B. Huang, M.~Ramesh, T.~Berg, and E.~Learned-Miller.
\newblock Labeled faces in the wild: A database for studying face recognition
  in unconstrained environments.
\newblock Technical report.

\bibitem{HubelWiesel}
D.~H. Hubel and T.~N. Wiesel.
\newblock Receptive fields, binocular interaction and functional architecture
  in the cat's visual cortex.
\newblock {\em The Journal of physiology}, 160(1):106--154, 1962.

\bibitem{Jafri:Survey09}
R.~Jafri and H.~R. Arabnia.
\newblock A survey of face recognition techniques.
\newblock 2009.

\bibitem{Jain:Biometrics}
A.~Jain, P.~Flynn, and A.~A. Ross.
\newblock {\em Handbook of Biometrics}.
\newblock Springer Science \& Business Media, 2007.

\bibitem{Jain:SoftBio}
A.~K. Jain, S.~C. Dass, and K.~Nandakumar.
\newblock Soft biometric traits for personal recognition systems.
\newblock In {\em Biometric Authentication}, pages 731--738. Springer, 2004.

\bibitem{Kass:IJCV88}
M.~Kass, A.~Witkin, and D.~Terzopoulos.
\newblock Snakes: Active contour models.
\newblock {\em International journal of computer vision}, 1(4):321--331, 1988.

\bibitem{Ira:CVPR14}
I.~Kemelmacher-Shlizerman, S.~Suwajanakorn, and S.~M. Seitz.
\newblock Illumination-aware age progression.
\newblock In {\em IEEE Conference on Computer Vision and Pattern Recognition
  (CVPR)}, pages 3334--3341, June 2014.

\bibitem{Kendall:84}
D.~G. Kendall.
\newblock Shape manifolds, procrustean metrics, and complex projective spaces.
\newblock {\em Bulletin of the London Mathematical Society}, 16(2):81--121,
  1984.

\bibitem{Kim:ICIP15}
J.~Kim, D.~Han, S.~Sohn, and J.~Kim.
\newblock Facial age estimation via extended curvature gabor filter.
\newblock In {\em IEEE International Conference on Image Processing (ICIP)},
  pages 1165--1169, Sept 2015.

\bibitem{Kohli:BTAS12}
N.~Kohli, R.~Singh, and M.~Vatsa.
\newblock Self-similarity representation of weber faces for kinship
  classification.
\newblock In {\em IEEE International Conference on Biometrics: Theory,
  Applications and Systems (BTAS)}, pages 245--250. IEEE, 2012.

\bibitem{Kohli:TIP16}
N.~Kohli, M.~Vatsa, R.~Singh, A.~Noore, and A.~Majumdar.
\newblock Hierarchical representation learning for kinship verification.
\newblock {\em IEEE Transactions on Image Processing}, 26(1):289--302, Jan
  2017.

\bibitem{Kong:TIP15}
S.~Kong, Z.~Jiang, and Q.~Yang.
\newblock Modeling neuron selectivity over simple midlevel features for image
  classification.
\newblock {\em IEEE Transactions on Image Processing}, 24(8):2404--2414, Aug
  2015.

\bibitem{Kuang:ICCVw15}
Z.~Kuang, C.~Huang, and W.~Zhang.
\newblock Deeply learned rich coding for cross-dataset facial age estimation.
\newblock In {\em IEEE International Conference on Computer Vision Workshop
  (ICCV-W)}, pages 338--343, Dec 2015.

\bibitem{Kwon:CVPR94}
Y.~H. Kwon et~al.
\newblock Age classification from facial images.
\newblock In {\em IEEE Conference on Computer Vision and Pattern Recognition
  (CVPR)}, pages 762--767. IEEE, 1994.

\bibitem{FGNET}
A.~Lanitis.
\newblock F{G-NET} {A}ging {D}atabase.
\newblock 2002.

\bibitem{Lanitis:EURASIP08}
A.~Lanitis.
\newblock Comparative evaluation of automatic age-progression methodologies.
\newblock {\em EURASIP Journal on Advances in Signal Processing}, 2008:101,
  2008.

\bibitem{Lanitis:FG00}
A.~Lanitis and C.~J. Taylor.
\newblock Towards automatic face identification robust to ageing variation.
\newblock In {\em IEEE International Conference and Workshops on Automatic Face
  and Gesture Recognition (FG)}, pages 391--396. IEEE, 2000.

\bibitem{Lanitis:TPAMI02prog}
A.~Lanitis, C.~J. Taylor, and T.~F. Cootes.
\newblock Toward automatic simulation of aging effects on face images.
\newblock {\em IEEE Transactions on Pattern Analysis and Machine Intelligence},
  24(4):442--455, 2002.

\bibitem{Lanitis:TPAMI02}
A.~Lanitis, C.~J. Taylor, and T.~F. Cootes.
\newblock Toward automatic simulation of aging effects on face images.
\newblock {\em IEEE Transactions on Pattern Analysis and Machine Intelligence},
  24(4):442--455, 2002.

\bibitem{Lanitis:FG15}
A.~Lanitis, N.~Tsapatsoulis, K.~Soteriou, D.~Kuwahara, and S.~Morishima.
\newblock Fg2015 age progression evaluation.
\newblock In {\em IEEE International Conference and Workshops on Automatic Face
  and Gesture Recognition (FG)}, volume~1, pages 1--6, May 2015.

\bibitem{Lecun:LeNet}
Y.~LeCun, L.~Bottou, Y.~Bengio, and P.~Haffner.
\newblock Gradient-based learning applied to document recognition.
\newblock {\em Proceedings of the IEEE}, 86(11):2278--2324, 1998.

\bibitem{Li:TransCyber15}
C.~Li, Q.~Liu, W.~Dong, X.~Zhu, J.~Liu, and H.~Lu.
\newblock Human age estimation based on locality and ordinal information.
\newblock {\em IEEE Transactions on Cybernetics}, 45(11):2522--2534, Nov 2015.

\bibitem{Li:CVPR12}
C.~Li, Q.~Liu, J.~Liu, and H.~Lu.
\newblock Learning ordinal discriminative features for age estimation.
\newblock In {\em IEEE Conference on Computer Vision and Pattern Recognition
  (CVPR)}, pages 2570--2577, June 2012.

\bibitem{Li:TIP16}
Z.~Li, D.~Gong, X.~Li, and D.~Tao.
\newblock Aging face recognition: A hierarchical learning model based on local
  patterns selection.
\newblock {\em IEEE Transactions on Image Processing}, 25(5):2146--2154, May
  2016.

\bibitem{Li:TIFS11}
Z.~Li, U.~Park, and A.~K. Jain.
\newblock A discriminative model for age invariant face recognition.
\newblock {\em IEEE transactions on information forensics and security},
  6(3):1028--1037, 2011.

\bibitem{Ling:ICCV07}
H.~Ling, S.~Soatto, N.~Ramanathan, and D.~W. Jacobs.
\newblock A study of face recognition as people age.
\newblock In {\em IEEE International Conference on Computer Vision (ICCV)},
  pages 1--8. IEEE, 2007.

\bibitem{Ling:TIFS10}
H.~Ling, S.~Soatto, N.~Ramanathan, and D.~W. Jacobs.
\newblock Face verification across age progression using discriminative
  methods.
\newblock {\em IEEE Transactions on Information Forensics and Security},
  5(1):82--91, March 2010.

\bibitem{Liu:TIP02}
C.~Liu and H.~Wechsler.
\newblock Gabor feature based classification using the enhanced fisher linear
  discriminant model for face recognition.
\newblock {\em IEEE Transactions on Image processing}, 11(4):467--476, 2002.

\bibitem{Liu:TPAMI11}
C.~Liu, J.~Yuen, and A.~Torralba.
\newblock Sift flow: Dense correspondence across scenes and its applications.
\newblock In {\em Dense Image Correspondences for Computer Vision}, pages
  15--49. Springer, 2016.

\bibitem{Liu:PR17}
H.~Liu, J.~Lu, J.~Feng, and J.~Zhou.
\newblock Group-aware deep feature learning for facial age estimation.
\newblock {\em Pattern Recognition}, 66:82--94, 2017.

\bibitem{Liu:TIFS17}
H.~Liu, J.~Lu, J.~Feng, and J.~Zhou.
\newblock Label-sensitive deep metric learning for facial age estimation.
\newblock {\em IEEE Transactions on Information Forensics and Security}, 2017.

\bibitem{Liu:TIFS18}
H.~Liu, J.~Lu, J.~Feng, and J.~Zhou.
\newblock Label-sensitive deep metric learning for facial age estimation.
\newblock {\em IEEE Transactions on Information Forensics and Security},
  13(2):292--305, 2018.

\bibitem{Liu:ICIP16}
H.~Liu and X.~Sun.
\newblock Linear canonical correlation analysis based ranking approach for
  facial age estimation.
\newblock In {\em IEEE International Conference on Image Processing (ICIP)},
  pages 3249--3253, Sept 2016.

\bibitem{Liu:ICASSP16}
H.~Liu and X.~Sun.
\newblock A partial least squares based ranker for fast and accurate age
  estimation.
\newblock In {\em IEEE International Conference on Acoustics, Speech and Signal
  Processing (ICASSP)}, pages 2792--2796. IEEE, 2016.

\bibitem{Liu:NIPS16}
M.-Y. Liu and O.~Tuzel.
\newblock Coupled generative adversarial networks.
\newblock In {\em Advances in neural information processing systems}, pages
  469--477, 2016.

\bibitem{Liu:BTAS15}
Q.~Liu, A.~Puthenputhussery, and C.~Liu.
\newblock Inheritable fisher vector feature for kinship verification.
\newblock In {\em IEEE International Conference on Biometrics: Theory,
  Applications and Systems (BTAS)}, pages 1--6. IEEE, 2015.

\bibitem{Liu:ICCVw15}
X.~Liu, S.~Li, M.~Kan, J.~Zhang, S.~Wu, W.~Liu, H.~Han, S.~Shan, and X.~Chen.
\newblock Agenet: Deeply learned regressor and classifier for robust apparent
  age estimation.
\newblock In {\em IEEE International Conference on Computer Vision Workshop
  (ICCV- W)}, pages 258--266, Dec 2015.

\bibitem{Lock:JIVE}
E.~F. Lock, K.~A. Hoadley, J.~S. Marron, and A.~B. Nobel.
\newblock Joint and individual variation explained (jive) for integrated
  analysis of multiple data types.
\newblock {\em The annals of applied statistics}, 7(1):523, 2013.

\bibitem{Lopez:TPAMI16}
M.~B. Lopez, E.~Boutellaa, and A.~Hadid.
\newblock Comments on the kinship face in the wild data sets.
\newblock {\em IEEE Transactions on Pattern Analysis and Machine Intelligence},
  38(11):2342--2344, Nov 2016.

\bibitem{Lowe:SIFT}
D.~G. Lowe.
\newblock Distinctive image features from scale-invariant keypoints.
\newblock {\em International journal of computer vision}, 60(2):91--110, 2004.

\bibitem{Lu:kincomp15}
J.~Lu, J.~Hu, V.~E. Liong, X.~Zhou, A.~Bottino, I.~U. Islam, T.~F. Vieira,
  X.~Qin, X.~Tan, S.~Chen, et~al.
\newblock The fg 2015 kinship verification in the wild evaluation.
\newblock In {\em IEEE International Conference and Workshops on Automatic Face
  and Gesture Recognition (FG)}, volume~1, pages 1--7. IEEE, 2015.

\bibitem{Lu:TIP17}
J.~Lu, J.~Hu, and Y.-P. Tan.
\newblock Discriminative deep metric learning for face and kinship
  verification.
\newblock {\em IEEE Transactions on Image Processing}, 2017.

\bibitem{Lu:kincomp14}
J.~Lu, J.~Hu, X.~Zhou, J.~Zhou, M.~Castrill{\'o}n-Santana, J.~Lorenzo-Navarro,
  L.~Kou, Y.~Shang, A.~Bottino, and T.~F. Vieira.
\newblock Kinship verification in the wild: The first kinship verification
  competition.
\newblock In {\em Biometrics (IJCB), 2014 IEEE International Joint Conference
  on}, pages 1--6. IEEE, 2014.

\bibitem{Lu:TIP15}
J.~Lu, V.~E. Liong, and J.~Zhou.
\newblock Cost-sensitive local binary feature learning for facial age
  estimation.
\newblock {\em IEEE Transactions on Image Processing}, 24(12):5356--5368, Dec
  2015.

\bibitem{Lu:TPAMI14}
J.~Lu, X.~Zhou, Y.-P. Tan, Y.~Shang, and J.~Zhou.
\newblock Neighborhood repulsed metric learning for kinship verification.
\newblock {\em IEEE transactions on pattern analysis and machine intelligence},
  36(2):331--345, 2014.

\bibitem{Mallat:GroupScatter}
S.~Mallat.
\newblock Group invariant scattering.
\newblock {\em Communications on Pure and Applied Mathematics},
  65(10):1331--1398, 2012.

\bibitem{Mallat:Wavelets}
S.~G. Mallat.
\newblock A theory for multiresolution signal decomposition: the wavelet
  representation.
\newblock {\em IEEE transactions on pattern analysis and machine intelligence},
  11(7):674--693, 1989.

\bibitem{Maronidis:VISAPP13}
A.~Maronidis and A.~Lanitis.
\newblock Facial age simulation using age-specific 3d models and recursive pca.
\newblock In {\em VISAPP (1)}, pages 663--668, 2013.

\bibitem{Mathias:ECCV14}
M.~Mathias, R.~Benenson, M.~Pedersoli, and L.~Van~Gool.
\newblock Face detection without bells and whistles.
\newblock In {\em European Conference on Computer Vision (ECCV)}, pages
  720--735. Springer, 2014.

\bibitem{Lifespan}
M.~Minear and D.~C. Park.
\newblock A lifespan database of adult facial stimuli.
\newblock {\em Behavior Research Methods}, 36(4):630--633, 2004.

\bibitem{Mirza:CGAN}
M.~Mirza and S.~Osindero.
\newblock Conditional generative adversarial nets.
\newblock {\em arXiv preprint arXiv:1411.1784}, 2014.

\bibitem{Moschoglou:CVPRW17}
S.~Moschoglou, A.~Papaioannou, C.~Sagonas, J.~Deng, I.~Kotsia, and
  S.~Zafeiriou.
\newblock Agedb: the first manually collected, in-the-wild age database.
\newblock In {\em IEEE Conference on Computer Vision and Pattern Recognition
  Workshop (CVPR- W)}, 2017.

\bibitem{Moyse:Age_est_psycho}
E.~Moyse.
\newblock Age estimation from faces and voices: a review.
\newblock {\em Psychologica Belgica}, 54(3), 2014.

\bibitem{Nanni:AIM10}
L.~Nanni, A.~Lumini, and S.~Brahnam.
\newblock Local binary patterns variants as texture descriptors for medical
  image analysis.
\newblock {\em Artificial intelligence in medicine}, 49(2):117--125, 2010.

\bibitem{Ng:Gender12}
C.~B. Ng, Y.~H. Tay, and B.~M. Goi.
\newblock Vision-based human gender recognition: A survey.
\newblock 2012.

\bibitem{Ni:ACM09}
B.~Ni, Z.~Song, and S.~Yan.
\newblock Web image mining towards universal age estimator.
\newblock In {\em Proceedings of the 17th ACM international conference on
  Multimedia}, pages 85--94. ACM, 2009.

\bibitem{Ni:TransMult11}
B.~Ni, Z.~Song, and S.~Yan.
\newblock Web image and video mining towards universal and robust age
  estimator.
\newblock {\em IEEE Transactions on Multimedia}, 13(6):1217--1229, Dec 2011.

\bibitem{Niu:CVPR16}
Z.~Niu, M.~Zhou, L.~Wang, X.~Gao, and G.~Hua.
\newblock Ordinal regression with multiple output cnn for age estimation.
\newblock In {\em The IEEE Conference on Computer Vision and Pattern
  Recognition (CVPR)}, June 2016.

\bibitem{BrownSisters}
G.~Nixon and P.~Galassi.
\newblock The {B}rown {S}isters: {T}hirty-{T}hree {Y}ears.
\newblock {\em Museum of Modern Art}, 2007.

\bibitem{Nixon:PRL15}
M.~S. Nixon, P.~L. Correia, K.~Nasrollahi, T.~B. Moeslund, A.~Hadid, and
  M.~Tistarelli.
\newblock On soft biometrics.
\newblock {\em Pattern Recognition Letters}, 68:218--230, 2015.

\bibitem{Ojala:LBP}
T.~Ojala, M.~Pietikainen, and T.~Maenpaa.
\newblock Multiresolution gray-scale and rotation invariant texture
  classification with local binary patterns.
\newblock {\em IEEE Transactions on pattern analysis and machine intelligence},
  24(7):971--987, 2002.

\bibitem{Ojansivu:ICISP08}
V.~Ojansivu and J.~Heikkil{\"a}.
\newblock Blur insensitive texture classification using local phase
  quantization.
\newblock In {\em International conference on image and signal processing},
  pages 236--243. Springer, 2008.

\bibitem{OToole:IMAVIS99}
A.~J. O'Toole, T.~Price, T.~Vetter, J.~Bartlett, and V.~Blanz.
\newblock 3d shape and 2d surface textures of human faces: the role of
  “averages” in attractiveness and age.
\newblock {\em Image and Vision Computing}, 18(1):9--19, 1999.

\bibitem{OToole:Perception97}
A.~J. O'toole, T.~Vetter, H.~Volz, and E.~M. Salter.
\newblock Three-dimensional caricatures of human heads: distinctiveness and the
  perception of facial age.
\newblock {\em Perception}, 26(6):719--732, 1997.

\bibitem{Pantic:TPAMI00}
M.~Pantic and L.~J.~M. Rothkrantz.
\newblock Automatic analysis of facial expressions: The state of the art.
\newblock {\em IEEE Transactions on pattern analysis and machine intelligence},
  22(12):1424--1445, 2000.

\bibitem{Park:2008}
J.~H. Park, M.~Schaller, and M.~Van~Vugt.
\newblock Psychology of human kin recognition: Heuristic cues, erroneous
  inferences, and their implications.
\newblock {\em Review of General Psychology}, 12(3):215, 2008.

\bibitem{Park:TPAMI10}
U.~Park, Y.~Tong, and A.~K. Jain.
\newblock Age-invariant face recognition.
\newblock {\em IEEE Trans. Pattern Anal. Mach. Intell.}, 32(5):947--954, May
  2010.

\bibitem{Patel:CVIU17}
B.~Patel, R.~Maheshwari, and B.~Raman.
\newblock Evaluation of periocular features for kinship verification in the
  wild.
\newblock {\em Computer Vision and Image Understanding}, 2017.

\bibitem{frgc}
P.~J. Phillips, P.~J. Flynn, T.~Scruggs, K.~W. Bowyer, J.~Chang, K.~Hoffman,
  J.~Marques, J.~Min, and W.~Worek.
\newblock Overview of the face recognition grand challenge.
\newblock In {\em IEEE Conference on Computer Vision and Pattern Recognition},
  volume~1, pages 947--954. IEEE, 2005.

\bibitem{feret}
P.~J. Phillips, H.~Wechsler, J.~Huang, and P.~J. Rauss.
\newblock The feret database and evaluation procedure for face-recognition
  algorithms.
\newblock {\em Image and vision computing}, 16(5):295--306, 1998.

\bibitem{Puthen:ICIP16}
A.~Puthenputhussery, Q.~Liu, and C.~Liu.
\newblock Sift flow based genetic fisher vector feature for kinship
  verification.
\newblock In {\em IEEE International Conference on Image Processing (ICIP)},
  pages 2921--2925, Sept 2016.

\bibitem{Qin:TransMulti15}
X.~Qin, X.~Tan, and S.~Chen.
\newblock Tri-subject kinship verification: Understanding the core of a family.
\newblock {\em IEEE Transactions on Multimedia}, 17(10):1855--1867, 2015.

\bibitem{Ramanathan:CVPR06}
N.~Ramanathan and R.~Chellappa.
\newblock Modeling age progression in young faces.
\newblock In {\em IEEE Computer Society Conference on Computer Vision and
  Pattern Recognition (CVPR)}, volume~1, pages 387--394, June 2006.

\bibitem{Ramanathan:FG08}
N.~Ramanathan and R.~Chellappa.
\newblock Modeling shape and textural variations in aging faces.
\newblock In {\em IEEE International Conference and Workshops on Automatic Face
  and Gesture Recognition (FG)}, pages 1--8. IEEE, 2008.

\bibitem{Ramanathan:JVLC09}
N.~Ramanathan, R.~Chellappa, and S.~Biswas.
\newblock Computational methods for modeling facial aging: A survey.
\newblock {\em Journal of Visual Languages \& Computing}, 20(3):131 -- 144,
  2009.

\bibitem{Ranjan:ICCVw15}
R.~Ranjan, S.~Zhou, J.~C. Chen, A.~Kumar, A.~Alavi, V.~M. Patel, and
  R.~Chellappa.
\newblock Unconstrained age estimation with deep convolutional neural networks.
\newblock In {\em IEEE International Conference on Computer Vision Workshop
  (ICCV- W)}, pages 351--359, Dec 2015.

\bibitem{Morph}
K.~Ricanek and T.~Tesafaye.
\newblock Morph: A longitudinal image database of normal adult age-progression.
\newblock In {\em IEEE International Conference on Automatic Face and Gesture
  Recognition (FG)}, pages 341--345. IEEE, 2006.

\bibitem{HMAX}
M.~Riesenhuber and T.~Poggio.
\newblock Hierarchical models of object recognition in cortex.
\newblock {\em Nature neuroscience}, 2(11):1019--1025, 1999.

\bibitem{Robinson:FIW}
J.~P. Robinson, M.~Shao, Y.~Wu, and Y.~Fu.
\newblock Family in the wild (fiw): A large-scale kinship recognition database.

\bibitem{FIW}
J.~P. Robinson, M.~Shao, Y.~Wu, and Y.~Fu.
\newblock Families in the wild (fiw): Large-scale kinship image database and
  benchmarks.
\newblock In {\em Proceedings of the 2016 ACM on Multimedia Conference}, MM
  '16, pages 242--246, New York, NY, USA, 2016. ACM.

\bibitem{Rothe:IJCV16}
R.~Rothe, R.~Timofte, and L.~V. Gool.
\newblock Deep expectation of real and apparent age from a single image without
  facial landmarks.
\newblock {\em International Journal of Computer Vision}, July 2016.

\bibitem{Russakovsky:Imagenet}
O.~Russakovsky, J.~Deng, H.~Su, J.~Krause, S.~Satheesh, S.~Ma, Z.~Huang,
  A.~Karpathy, A.~Khosla, M.~Bernstein, et~al.
\newblock Imagenet large scale visual recognition challenge.
\newblock {\em International Journal of Computer Vision}, 115(3):211--252,
  2015.

\bibitem{Sagonas:ICPR16}
C.~Sagonas, Y.~Panagakis, S.~Arunkumar, N.~Ratha, and S.~Zafeiriou.
\newblock Back to the future: A fully automatic method for robust age
  progression.
\newblock In {\em International Conference on Pattern Recognition (ICPR)},
  pages 4226--4231, Dec 2016.

\bibitem{Sagonas:CVPR17}
C.~Sagonas, Y.~Panagakis, A.~Leidinger, S.~Zafeiriou, et~al.
\newblock Robust joint and individual variance explained.
\newblock In {\em IEEE Computer Society Conference on Computer Vision and
  Pattern Recognition (CVPR)}, 2017.

\bibitem{Sala:DBM}
R.~Salakhutdinov and G.~Hinton.
\newblock Deep boltzmann machines.
\newblock In {\em Artificial Intelligence and Statistics}, pages 448--455,
  2009.

\bibitem{Sariyanidi:TPAMI15}
E.~Sariyanidi, H.~Gunes, and A.~Cavallaro.
\newblock Automatic analysis of facial affect: A survey of registration,
  representation, and recognition.
\newblock {\em IEEE transactions on pattern analysis and machine intelligence},
  37(6):1113--1133, 2015.

\bibitem{Schapire:99}
R.~E. Schapire and Y.~Singer.
\newblock Improved boosting algorithms using confidence-rated predictions.
\newblock {\em Machine learning}, 37(3):297--336, 1999.

\bibitem{Scherbaum:CGF07}
K.~Scherbaum, M.~Sunkel, H.-P. Seidel, and V.~Blanz.
\newblock Prediction of individual non-linear aging trajectories of faces.
\newblock In {\em Computer Graphics Forum}, volume~26, pages 285--294. Wiley
  Online Library, 2007.

\bibitem{Seung:Manifold}
H.~S. Seung and D.~D. Lee.
\newblock The manifold ways of perception.
\newblock {\em science}, 290(5500):2268--2269, 2000.

\bibitem{Shao:CVPRw11}
M.~Shao, S.~Xia, and Y.~Fu.
\newblock Genealogical face recognition based on ub kinface database.
\newblock In {\em IEEE Conference on Computer Vision and Pattern Recognition
  Workshop (CVPR- W)}, pages 60--65. IEEE, 2011.

\bibitem{Shechtman:CVPR07}
E.~Shechtman and M.~Irani.
\newblock Matching local self-similarities across images and videos.
\newblock In {\em IEEE Conference on Computer Vision and Pattern Recognition
  (CVPR)}. IEEE, 2007.

\bibitem{Shen:JISE14}
C.-T. Shen, F.~Huang, W.-H. Lu, S.-W. Shih, and H.-Y.~M. Liao.
\newblock 3d age progression prediction in children's faces with a small
  exemplar-image set.
\newblock {\em J. Inf. Sci. Eng.}, 30(4):1131--1148, 2014.

\bibitem{Shen:ISM11}
C.-T. Shen, W.-H. Lu, S.-W. Shih, and H.-Y.~M. Liao.
\newblock Exemplar-based age progression prediction in children faces.
\newblock In {\em IEEE International Symposium on Multimedia (ISM)}, pages
  123--128. IEEE, 2011.

\bibitem{Shu:ICCV15}
X.~Shu, J.~Tang, H.~Lai, L.~Liu, and S.~Yan.
\newblock Personalized age progression with aging dictionary.
\newblock In {\em IEEE International Conference on Computer Vision (ICCV)},
  pages 3970--3978, Dec 2015.

\bibitem{Shu:PR16}
X.~Shu, J.~Tang, H.~Lai, Z.~Niu, and S.~Yan.
\newblock Kinship-guided age progression.
\newblock {\em Pattern Recognition}, 59:156--167, 2016.

\bibitem{pie}
T.~Sim, S.~Baker, and M.~Bsat.
\newblock The cmu pose, illumination, and expression (pie) database.
\newblock In {\em IEEE International Conference on Automatic Face and Gesture
  Recognition (FG)}, pages 53--58. IEEE, 2002.

\bibitem{Simonyan:BMVC13}
K.~Simonyan, O.~M. Parkhi, A.~Vedaldi, and A.~Zisserman.
\newblock Fisher vector faces in the wild.

\bibitem{Simonyan:VGG16}
K.~Simonyan and A.~Zisserman.
\newblock Very deep convolutional networks for large-scale image recognition.
\newblock {\em arXiv preprint arXiv:1409.1556}, 2014.

\bibitem{Somanath:BTAS12}
G.~Somanath and C.~Kambhamettu.
\newblock Can faces verify blood-relations?
\newblock In {\em IEEE International Conference on Biometrics: Theory,
  Applications and Systems (BTAS)}, pages 105--112. IEEE, 2012.

\bibitem{Somanath:ECCV12}
G.~Somanath, M.~Rohith, and C.~Kambhamettu.
\newblock Vadana: A dense dataset for facial image analysis.
\newblock In {\em IEEE International Conference on Computer Vision Workshops
  (ICCV- W)}, pages 2175--2182. IEEE, 2011.

\bibitem{Song:ICCV11}
Z.~Song, B.~Ni, D.~Guo, T.~Sim, and S.~Yan.
\newblock Learning universal multi-view age estimator using video context.
\newblock In {\em IEEE International Conference on Computer Vision (ICCV)},
  pages 241--248, Nov 2011.

\bibitem{Stanley:CPNN}
K.~O. Stanley.
\newblock Compositional pattern producing networks: A novel abstraction of
  development.
\newblock {\em Genetic programming and evolvable machines}, 8(2):131--162,
  2007.

\bibitem{Su:ICASSP10}
Y.~Su, Y.~Fu, Q.~Tian, and X.~Gao.
\newblock Cross-database age estimation based on transfer learning.
\newblock In {\em IEEE International Conference on Acoustics, Speech and Signal
  Processing (ICASSP)}, pages 1270--1273, March 2010.

\bibitem{Suo:TPAMI12}
J.~Suo, X.~Chen, S.~Shan, W.~Gao, and Q.~Dai.
\newblock A concatenational graph evolution aging model.
\newblock {\em IEEE Transactions on Pattern Analysis and Machine Intelligence},
  34(11):2083--2096, Nov 2012.

\bibitem{Suo:TPAMI10}
J.~Suo, S.~C. Zhu, S.~Shan, and X.~Chen.
\newblock A compositional and dynamic model for face aging.
\newblock {\em IEEE Transactions on Pattern Analysis and Machine Intelligence},
  32(3):385--401, March 2010.

\bibitem{Szegedy:GoogLeNet}
C.~Szegedy, W.~Liu, Y.~Jia, P.~Sermanet, S.~Reed, D.~Anguelov, D.~Erhan,
  V.~Vanhoucke, and A.~Rabinovich.
\newblock Going deeper with convolutions.
\newblock In {\em IEEE Conference on Computer Vision and Pattern Recognition
  (CVPR)}, pages 1--9, 2015.

\bibitem{Tan:LTP}
X.~Tan and B.~Triggs.
\newblock Enhanced local texture feature sets for face recognition under
  difficult lighting conditions.
\newblock {\em IEEE transactions on image processing}, 19(6):1635--1650, 2010.

\bibitem{Thukral:ICASSP12}
P.~Thukral, K.~Mitra, and R.~Chellappa.
\newblock A hierarchical approach for human age estimation.
\newblock In {\em 2012 IEEE International Conference on Acoustics, Speech and
  Signal Processing (ICASSP)}, pages 1529--1532. IEEE, 2012.

\bibitem{Tiddeman:CGA01}
B.~Tiddeman, M.~Burt, and D.~Perrett.
\newblock Prototyping and transforming facial textures for perception research.
\newblock {\em IEEE computer graphics and applications}, 21(5):42--50, 2001.

\bibitem{Tipping:JMLR01}
M.~E. Tipping.
\newblock Sparse bayesian learning and the relevance vector machine.
\newblock {\em Journal of machine learning research}, 1(Jun):211--244, 2001.

\bibitem{Tola:TPAMI10}
E.~Tola, V.~Lepetit, and P.~Fua.
\newblock Daisy: An efficient dense descriptor applied to wide-baseline stereo.
\newblock {\em IEEE transactions on pattern analysis and machine intelligence},
  32(5):815--830, 2010.

\bibitem{Turk:Eigenfaces}
M.~A. Turk and A.~P. Pentland.
\newblock Face recognition using eigenfaces.
\newblock In {\em IEEE Conference on Computer Vision and Pattern Recognition
  (CVPR)}, pages 586--591. IEEE, 1991.

\bibitem{witbd}
K.~Ueki, T.~Hayashida, and T.~Kobayashi.
\newblock Subspace-based age-group classification using facial images under
  various lighting conditions.
\newblock In {\em Automatic Face and Gesture Recognition, 2006. FGR 2006. 7th
  International Conference on}, pages 6--pp. IEEE, 2006.

\bibitem{Uricar:CVPRw16}
M.~Uricar, R.~Timofte, R.~Rothe, J.~Matas, and L.~Van~Gool.
\newblock Structured output svm prediction of apparent age, gender and smile
  from deep features.
\newblock In {\em The IEEE Conference on Computer Vision and Pattern
  Recognition (CVPR) Workshops}, June 2016.

\bibitem{Vapnik:LUPI}
V.~Vapnik and A.~Vashist.
\newblock A new learning paradigm: Learning using privileged information.
\newblock {\em Neural networks}, 22(5):544--557, 2009.

\bibitem{Vapnik:SVM}
V.~N. Vapnik and V.~Vapnik.
\newblock {\em Statistical learning theory}, volume~1.
\newblock Wiley New York, 1998.

\bibitem{Vieira:VisCom14}
T.~F. Vieira, A.~Bottino, A.~Laurentini, and M.~De~Simone.
\newblock Detecting siblings in image pairs.
\newblock {\em The Visual Computer}, 30(12):1333--1345, 2014.

\bibitem{ViolaJones}
P.~Viola and M.~J. Jones.
\newblock Robust real-time face detection.
\newblock {\em International journal of computer vision}, 57(2):137--154, 2004.

\bibitem{Wang:PR17}
N.~Wang, X.~Gao, D.~Tao, H.~Yang, and X.~Li.
\newblock Facial feature point detection: A comprehensive survey.
\newblock {\em Neurocomputing}, 2017.

\bibitem{Wang:TCyber15}
S.~Wang, D.~Tao, and J.~Yang.
\newblock Relative attribute svm+ for learning for age estimation.
\newblock {\em IEEE Transactions on Cybernetics}, 46(3):827--839, March 2016.

\bibitem{Wang:CVPR16}
W.~Wang, Z.~Cui, Y.~Yan, J.~Feng, S.~Yan, X.~Shu, and N.~Sebe.
\newblock Recurrent face aging.
\newblock In {\em The IEEE Conference on Computer Vision and Pattern
  Recognition (CVPR)}, June 2016.

\bibitem{Wang:ICIP14}
X.~Wang and C.~Kambhamettu.
\newblock Leveraging appearance and geometry for kinship verification.
\newblock In {\em IEEE International Conference on Image Processing (ICIP)},
  pages 5017--5021. IEEE, 2014.

\bibitem{Wang:TSMC12}
Y.~Wang, Z.~Zhang, W.~Li, and F.~Jiang.
\newblock Combining tensor space analysis and active appearance models for
  aging effect simulation on face images.
\newblock {\em IEEE Transactions on Systems, Man, and Cybernetics, Part B
  (Cybernetics)}, 42(4):1107--1118, Aug 2012.

\bibitem{Wen:CVPR16}
Y.~Wen, Z.~Li, and Y.~Qiao.
\newblock Latent factor guided convolutional neural networks for age-invariant
  face recognition.
\newblock In {\em The IEEE Conference on Computer Vision and Pattern
  Recognition (CVPR)}, June 2016.

\bibitem{Wolf:ECCVw08}
L.~Wolf, T.~Hassner, and Y.~Taigman.
\newblock Descriptor based methods in the wild.
\newblock In {\em European Conference on Computer Vision Workshop (ECCV- W)},
  2008.

\bibitem{Wu:TIFS12}
T.~Wu, P.~Turaga, and R.~Chellappa.
\newblock Age estimation and face verification across aging using landmarks.
\newblock {\em IEEE Transactions on Information Forensics and Security},
  7(6):1780--1788, Dec 2012.

\bibitem{Wu:99}
Y.~Wu, P.~Kalra, L.~Moccozet, and N.~Magnenat-Thalmann.
\newblock Simulating wrinkles and skin aging.
\newblock {\em The visual computer}, 15(4):183--198, 1999.

\bibitem{Wu:94}
Y.~Wu, N.~M. Thalmann, and D.~Thalmann.
\newblock A plastic-visco-elastic model for wrinkles in facial animation and
  skin aging.

\bibitem{Wu:95}
Y.~Wu, N.~M. Thalmann, and D.~Thalmann.
\newblock A dynamic wrinkle model in facial animation and skin ageing.
\newblock {\em Computer Animation and Virtual Worlds}, 6(4):195--205, 1995.

\bibitem{Xia:VISAPP14}
B.~Xia, B.~B. Amor, M.~Daoudi, and H.~Drira.
\newblock Can 3d shape of the face reveal your age?
\newblock In {\em 2014 International Conference on Computer Vision Theory and
  Applications (VISAPP)}, volume~2, pages 5--13, Jan 2014.

\bibitem{xia:ub1}
S.~Xia, M.~Shao, and Y.~Fu.
\newblock Kinship verification through transfer learning.

\bibitem{Xia:IJCAI11}
S.~Xia, M.~Shao, and Y.~Fu.
\newblock Kinship verification through transfer learning.
\newblock In {\em IJCAI Proceedings-international joint conference on
  artificial intelligence}, volume~22, page 2539, 2011.

\bibitem{Xia:ICPR12}
S.~Xia, M.~Shao, and Y.~Fu.
\newblock Toward kinship verification using visual attributes.
\newblock In {\em International Conference on Pattern Recognition (ICPR)},
  pages 549--552. IEEE, 2012.

\bibitem{Xia:TransMulti12}
S.~Xia, M.~Shao, J.~Luo, and Y.~Fu.
\newblock Understanding kin relationships in a photo.
\newblock {\em IEEE Transactions on Multimedia}, 14(4):1046--1056, 2012.

\bibitem{Xing:MetricLearning}
E.~P. Xing, M.~I. Jordan, S.~J. Russell, and A.~Y. Ng.
\newblock Distance metric learning with application to clustering with
  side-information.
\newblock In {\em Advances in neural information processing systems}, pages
  521--528, 2003.

\bibitem{Xing:PR17}
J.~Xing, K.~Li, W.~Hu, C.~Yuan, and H.~Ling.
\newblock Diagnosing deep learning models for high accuracy age estimation from
  a single image.
\newblock {\em Pattern Recognition}, 66:106--116, 2017.

\bibitem{Yan:IMAVIS17}
H.~Yan.
\newblock Kinship verification using neighborhood repulsed correlation metric
  learning.
\newblock {\em Image and Vision Computing}, 60:91--97, 2017.

\bibitem{Yan:PR17}
H.~Yan and J.~Hu.
\newblock Video-based kinship verification using distance metric learning.
\newblock {\em Pattern Recognition}, 2017.

\bibitem{Yan:TIFS14}
H.~Yan, J.~Lu, W.~Deng, and X.~Zhou.
\newblock Discriminative multimetric learning for kinship verification.
\newblock {\em IEEE Transactions on Information forensics and security},
  9(7):1169--1178, 2014.

\bibitem{Yan:TransCYber15}
H.~Yan, J.~Lu, and X.~Zhou.
\newblock Prototype-based discriminative feature learning for kinship
  verification.
\newblock {\em IEEE transactions on cybernetics}, 45(11):2535--2545, 2015.

\bibitem{Yang:TIP16}
H.~Yang, D.~Huang, Y.~Wang, H.~Wang, and Y.~Tang.
\newblock Face aging effect simulation using hidden factor analysis joint
  sparse representation.
\newblock {\em IEEE Transactions on Image Processing}, 25(6):2493--2507, 2016.

\bibitem{Yang:CVPR11}
M.~Yang, S.~Zhu, F.~Lv, and K.~Yu.
\newblock Correspondence driven adaptation for human profile recognition.
\newblock In {\em IEEE Conference on Computer Vision and Pattern Recognition
  (CVPR)}, pages 505--512, June 2011.

\bibitem{Yang:Bio07}
Z.~Yang and H.~Ai.
\newblock Demographic classification with local binary patterns.
\newblock {\em Advances in Biometrics}, pages 464--473, 2007.

\bibitem{Yi:ACCV14}
D.~Yi, Z.~Lei, and S.~Z. Li.
\newblock Age estimation by multi-scale convolutional network.
\newblock In {\em Asian Conference on Computer Vision (ACCV)}, pages 144--158.
  Springer, 2014.

\bibitem{Zafeiriou:CVIU15}
S.~Zafeiriou, C.~Zhang, and Z.~Zhang.
\newblock A survey on face detection in the wild: past, present and future.
\newblock {\em Computer Vision and Image Understanding}, 138:1--24, 2015.

\bibitem{Zebrowitz:2008}
L.~A. Zebrowitz and J.~M. Montepare.
\newblock Social psychological face perception: Why appearance matters.
\newblock {\em Social and Personality Psychology Compass}, 2(3):1497--1517,
  2008.

\bibitem{Zeng:TPAMI09}
Z.~Zeng, M.~Pantic, G.~I. Roisman, and T.~S. Huang.
\newblock A survey of affect recognition methods: Audio, visual, and
  spontaneous expressions.
\newblock {\em IEEE transactions on pattern analysis and machine intelligence},
  31(1):39--58, 2009.

\bibitem{Zhang:TIP07}
B.~Zhang, S.~Shan, X.~Chen, and W.~Gao.
\newblock Histogram of gabor phase patterns (hgpp): A novel object
  representation approach for face recognition.
\newblock {\em IEEE Transactions on Image Processing}, 16(1):57--68, 2007.

\bibitem{Zhang:ICIP16}
J.~Zhang, S.~Xia, H.~Pan, and A.~Qin.
\newblock A genetics-motivated unsupervised model for tri-subject kinship
  verification.
\newblock In {\em IEEE International Conference on Image Processing (ICIP)},
  pages 2916--2920. IEEE, 2016.

\bibitem{Zhang:SPL16}
K.~Zhang, Z.~Zhang, Z.~Li, and Y.~Qiao.
\newblock Joint face detection and alignment using multitask cascaded
  convolutional networks.
\newblock {\em IEEE Signal Processing Letters}, 23(10):1499--1503, 2016.

\bibitem{Zhang:TPAMI15}
M.-L. Zhang and L.~Wu.
\newblock Lift: Multi-label learning with label-specific features.
\newblock {\em IEEE transactions on pattern analysis and machine intelligence},
  37(1):107--120, 2015.

\bibitem{Zhang:CVPR10}
Y.~Zhang and D.~Y. Yeung.
\newblock Multi-task warped gaussian process for personalized age estimation.
\newblock In {\em 2010 IEEE Computer Society Conference on Computer Vision and
  Pattern Recognition}, pages 2622--2629, June 2010.

\bibitem{Zhang:CVPR17}
Z.~Zhang, Y.~Song, and H.~Qi.
\newblock Age progression/regression by conditional adversarial autoencoder.
\newblock In {\em CVPR}, 2017.

\bibitem{Zhao:EBGAN}
J.~Zhao, M.~Mathieu, and Y.~LeCun.
\newblock Energy-based generative adversarial networks.
\newblock In {\em International Conference on Learning Representations (ICLR)},
  2017.

\bibitem{Zhao:03}
W.~Zhao, R.~Chellappa, P.~J. Phillips, and A.~Rosenfeld.
\newblock Face recognition: A literature survey.
\newblock {\em ACM computing surveys (CSUR)}, 35(4):399--458, 2003.

\bibitem{Zhou:ICCV05}
S.~K. Zhou, B.~Georgescu, X.~S. Zhou, and D.~Comaniciu.
\newblock Image based regression using boosting method.
\newblock In {\em IEEE International Conference on Computer Vision (ICCV)},
  volume~1, pages 541--548. IEEE, 2005.

\bibitem{Zhou:ACM11}
X.~Zhou, J.~Hu, J.~Lu, Y.~Shang, and Y.~Guan.
\newblock Kinship verification from facial images under uncontrolled
  conditions.
\newblock In {\em Proceedings of the 19th ACM international conference on
  Multimedia}, pages 953--956. ACM, 2011.

\bibitem{Zhu:ICCVw15}
Y.~Zhu, Y.~Li, G.~Mu, and G.~Guo.
\newblock A study on apparent age estimation.
\newblock In {\em IEEE International Conference on Computer Vision Workshop
  (ICCV-W)}, pages 267--273, Dec 2015.

\end{thebibliography}
}}

%\comments{

\end{document}